\documentclass[10pt,journal,compsoc]{IEEEtran}
\usepackage{amsmath,amsfonts}
\usepackage{lipsum} % for dummy text
\usepackage{algorithmic}
\usepackage{algorithm}
\usepackage{array}
\usepackage{color}
\usepackage[caption=false,font=normalsize,labelfont=sf,textfont=sf]{subfig}
\usepackage{textcomp}
\usepackage{url}
\usepackage{verbatim}
\usepackage{graphicx}
\usepackage{cite}
\usepackage{longtable}
\usepackage[table,xcdraw]{xcolor}
\usepackage{amssymb}
\usepackage{graphicx}
\usepackage{pifont}

\usepackage[table]{xcolor}
\usepackage{}
\usepackage{tabularx}
\usepackage[table]{xcolor}
\usepackage{makecell}
\definecolor{risklow}{RGB}{198,239,206}     % Green
\definecolor{riskmed}{RGB}{255,235,156}     % Yellow
\definecolor{riskhigh}{RGB}{255,199,206}    % Red
\usepackage{enumitem}  

\usepackage{hyperref}

\begin{document}

\title{A Survey on Autonomy-Induced Security Risks in Large Model-Based Agents}

\author{Hang Su, Jun Luo, Chang Liu, Xiao Yang, Yichi Zhang, Yinpeng Dong, Jun Zhu~\IEEEmembership{Fellow,~IEEE}
        % <-this % stops a space
\thanks{
Hang Su, Jun Luo, Chang Liu, Xiao Yang, Yichi Zhang, Yinpeng Dong, and Jun Zhu are with Dept. of Comp. Sci. \& Tech., College of AI, Institute for AI, BNRist Center, THBI Lab, Tsinghua-Bosch Joint Center for ML, Tsinghua University, Beijing, China; Email: suhangss@mail.tsinghua.edu.cn,
luojunlymf@gmail.com, liuchang6513@tsinghua.edu.cn, yangxiao19@tsinghua.org.cn, \{zyc22@mails., dongyinpeng,  dcszj\}@tsinghua.edu.cn.}
\thanks{Corresponding author: Hang Su, Jun Zhu}

}

\markboth{IEEE TRANSACTIONS ON PATTERN ANALYSIS AND MACHINE INTELLIGENCE}%
{Shell \MakeLowercase{\textit{et al.}}: A Sample Article Using IEEEtran.cls for IEEE Journals}

\IEEEtitleabstractindextext{%
\begin{abstract}

Recent advances in large language models (LLMs) have catalyzed the rise of autonomous AI agents capable of perceiving, reasoning, and acting in dynamic, open-ended environments. These large-model agents mark a paradigm shift from static inference systems to interactive, memory-augmented entities. While these capabilities significantly expand the functional scope of AI, they also introduce qualitatively novel security risks—such as memory poisoning, tool misuse, reward hacking, and emergent misalignment—that extend beyond the threat models of conventional systems or standalone LLMs. In this survey, we first examine the structural foundations and key capabilities that underpin increasing levels of agent autonomy, including long-term memory retention, modular tool use, recursive planning, and reflective reasoning. We then analyze the corresponding security vulnerabilities across the agent stack, identifying failure modes such as deferred decision hazards, irreversible tool chains, and deceptive behaviors arising from internal state drift or value misalignment. These risks are traced to architectural fragilities that emerge across perception, cognition, memory, and action modules. To address these challenges, we systematically review recent defense strategies deployed at different autonomy layers, including input sanitization, memory lifecycle control, constrained decision-making, structured tool invocation, and introspective reflection. While these techniques provide partial mitigation, most operate in isolation and lack the integrated coherence required to manage emergent, temporally extended, and cross-module threats. Motivated by these limitations, we introduce the Reflective Risk-Aware Agent Architecture (R2A2)—a unified cognitive framework grounded in Constrained Markov Decision Processes (CMDPs), which incorporates risk-aware world modeling, meta-policy adaptation, and joint reward–risk optimization to enable principled, proactive safety across the agent's decision-making loop. This survey provides a structured understanding of how autonomy reshapes the security landscape of intelligent systems and offers a blueprint for embedding safety as a core design principle in next-generation AI agents.
\end{abstract}


}

\maketitle


\IEEEdisplaynontitleabstractindextext


% \fi
%\IEEEpeerreviewmaketitle

\section{Introduction}

\label{Introduction}
%\subsection{Background}

Recent advances in artificial intelligence have given rise to a new class of autonomous AI agents\cite{sapkota2025ai,Xi2023Rise} built on large-scale models\cite{bommasani2021opportunities,openai2023gpt,grattafiori2024llama}. Unlike traditional AI systems that output a single prediction or decision for a given input, these \textbf{large model AI agents} (often powered by state-of-the-art large language models, LLMs)\cite{yao2023react,schick2023toolformer} can interact continually with their environment. They perceive inputs (from users or other sources), reason about what to do next, and then act through various tools\cite{Nakano2021WebGPT} or actuators–all in a closed feedback loop. Early prototypes such as interactive chatbots with tool access have demonstrated that an LLM augmented with memory\cite{wang2023voyager} and the ability to execute commands can perform multi-step tasks without constant human oversight\cite{yao2023react,schick2023toolformer,yang2024AutoGPT}. This marks a significant shift from viewing AI as static models to treating them as active, situated agents, blurring the line between software and robot in cyberspace~\cite{he2024security}. The implications of this shift are profound, especially in terms of security, where the autonomy and broad capabilities of such agents introduce both new opportunities and new risks.

% \begin{figure}[!tbp]
%     \centering
%    % \includegraphics[width=0.45\textwidth]{figures/framework.pdf}
%       \includegraphics[width=0.98\linewidth]{figures/framework_en.pdf}
%     \caption{A conceptual architecture of an LLM-powered AI agent interacting with its environment. The agent (dashed box) integrates a large language model (LLM) as the core reasoning engine, a suite of external tools/APIs for acting on the world (e.g. web services, databases, operating system commands), and a memory store for long-term knowledge. }
%     \label{fig:agent_framework}
% \end{figure}

In each cycle, the agent receives an input (user query or environmental feedback) and feeds it to the LLM, which produces an action or decision. The action may involve calling a tool (e.g., querying a database or executing code), after which the output of the tool is fed back into the agent as new information~\cite{schick2023toolformer, yao2023react, wu2023autogen,yang2024AutoGPT}. This perception–action loop enables the agent to operate autonomously: \emph{it can refine its plans based on intermediate results, pursue goals through multiple steps, and even update its internal memory with new data. In essence, large-model agents \textbf{transform static AI models into adaptive decision-makers} that can continuously learn from and influence their environment. }

\begin{figure*}[!t]
    \centering
    \includegraphics[width=1\linewidth]{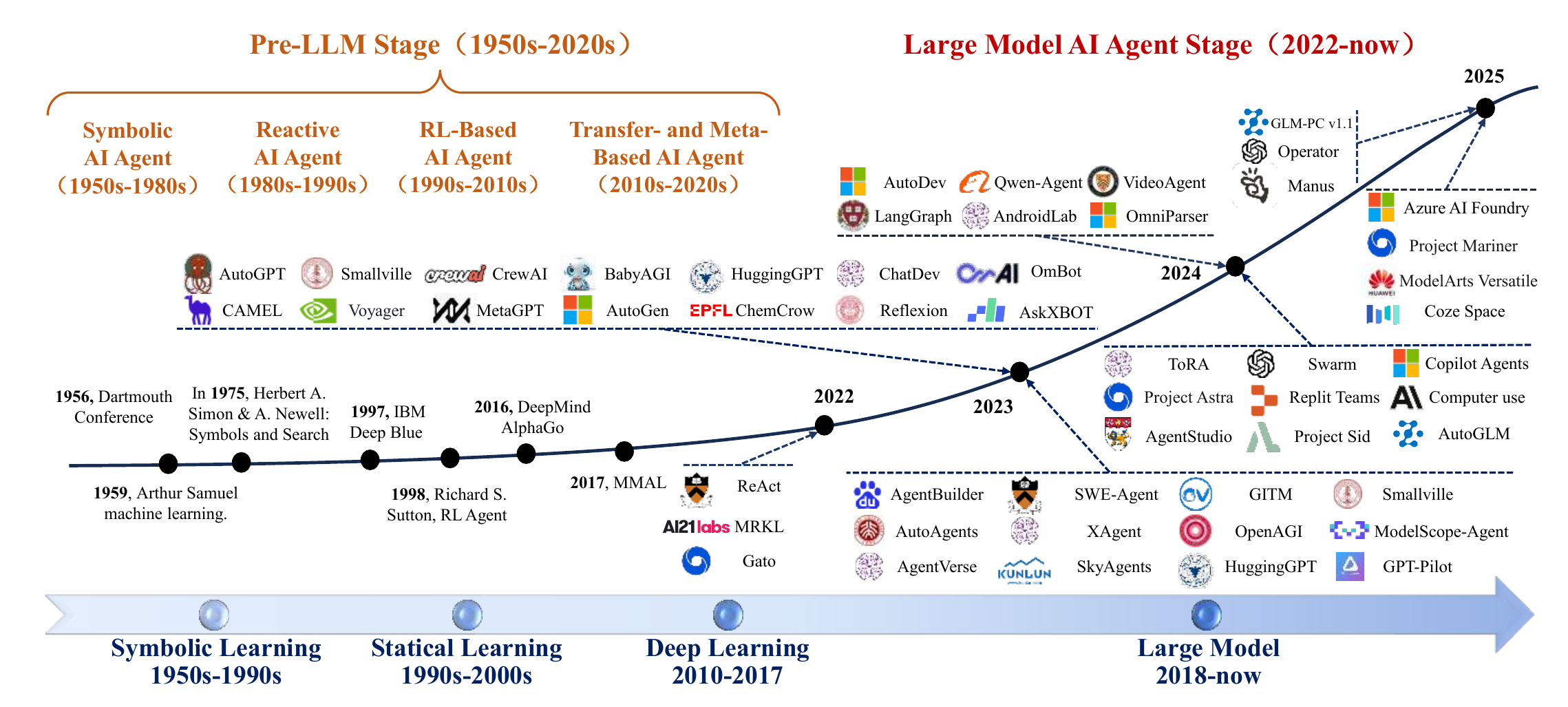}
    \vspace{-5ex}
      \caption{Historical evolution of AI agent paradigms. The development trajectory is categorized into the Pre-LLM Stage (1950s–2020s), encompassing symbolic, reactive, reinforcement learning (RL)-based, and meta/transfer-based agents, and the Large Model AI Agent Stage (2022–present), marked by the emergence of agents powered by large foundation models. Key representative projects and institutions are aligned along their respective paradigms, highlighting the shift from rule-based systems to tool-using, memory-equipped, and planning-capable LLM agents. }
         \vspace{-2ex}
    \label{Agent_trend}
\end{figure*}

This paradigm change from static models to autonomous LLM-driven agents is reminiscent of a leap from single-turn to ongoing intelligence as shown in Figure~\ref{Agent_trend}. Traditional AI agents (for example, early expert systems or fixed-rule bots) operated on predefined rules or narrow models and thus struggled to generalize beyond their programmed scope\cite{russell2021ai,dunzelman2023comprehensive}.
In contrast, an LLM-based agent inherits the open-ended problem-solving capacity of its underlying model, giving it a much wider action space. It can read documentation or dynamic context at runtime and figure out how to use new tools on the fly\cite{wang2023survey}. This flexibility has fueled excitement that such agents could serve as broadly capable assistants in society, tackling complex tasks in diverse domains. Indeed, LLM agents are already being applied in areas like software coding, web automation, personal assistance, and even embodied tasks such as robotic control, marking a significant step toward more general AI behavior~\cite{luo2025survey}. 

The property that has given humans a dominant advantage over other species is not strength or speed, but intelligence. If AI development continues its current trajectory, systems may eventually exceed human-level reasoning abilities across nearly all fields~\cite{bostrom2014superintelligence}. 
In fact, Bengio et al. warn that our pace of progress has outstripped safety efforts, and call for proactive risk management across multiple dimensions~\cite{bengio2024extremeAIRisks}. Such \emph{superintelligent} agents would possess the capacity to develop novel tools and strategies for controlling their environment~\cite{muehlhauser2012intelligence}. Unlike humans, they would not inherit our evolutionary instincts or motivations—yet most goals, even benign ones, are more efficiently pursued with increased resources~\cite{omohundro2018basic}. This default incentive structure could place their objectives in tension with human interests, potentially leading to deceptive, manipulative, or intervention-resistant behaviors~\cite{bostrom2014superintelligence}. 
To address these risks, Bengio et al. recently proposed a new paradigm known as \emph{Scientific AI}~\cite{bengio2025scientistai}, which emphasizes \emph{understanding before acting}. Rather than directly optimizing goals through unconstrained actions, Scientific AI agents prioritize constructing accurate, interpretable world models, generating causal hypotheses, and reasoning under uncertainty. This approach promotes introspection, modular reasoning, and verifiability, thereby reducing the risk of misaligned behavior~\cite{bengio2024aisafetyreport}.
As such, care must be taken to ensure the development of aligned agents—those which reliably pursue beneficial goals, cooperate with human oversight\cite{Borghoff2025Human}, and tolerate design imperfections. These foundational challenges—goal alignment, formal value specification, and corrigibility—constitute the core of long-term AI safety research~\cite{soares2017agent}.

Recent advances in large language models (LLMs)~\cite{wang2023describe} have given rise to a new generation of autonomous agents equipped with long-horizon planning, persistent memory, and the ability to invoke external tools. While these agents offer transformative potential across domains, their elevated autonomy introduces fundamentally new security challenges. Unlike static LLMs that passively generate text, autonomous agents can take consequential real-world actions---executing code, modifying databases, or calling APIs---thereby amplifying the risks associated with system failures and adversarial interventions. As summarized in Table~\ref{table:security_comparison_fused}, these threats stem directly from the very capabilities that empower such agents.
Specifically, multi-step reasoning, dynamic tool use, and environment-aware adaptation expand the attack surface across multiple architectural layers~\cite{deng2024aiagent,zhang2025advlm,zhang2024agent,li2025asb,zhang2024breaking,Hua2024TrustAgent,yu2025survey,li2025commercial,chen2025obvious,narajala2025securing}. The underlying LLM remains vulnerable to adversarial prompts and hallucinations~\cite{cheng2024exploring}; memory systems can be poisoned, manipulated, or exfiltrated; tool-use interfaces introduce execution pathways for unsafe behaviors; and planning modules may generate brittle action sequences or pursue misaligned objectives. Furthermore, these risks are exacerbated by agents’ interaction with open-ended, unpredictable environments---such as unverified web content or untrusted user inputs---which challenge traditional safety assumptions~\cite{yang2025mla,openai2024alignment,ma2024caution}.
\begin{table*}[t]
\centering
\caption{Escalating Security Risk Levels Across AI System Types with Representative Threats}
\label{table:security_comparison_fused}
\vspace{-1ex}
\renewcommand{\arraystretch}{1.4}
\begin{tabular}{|>{\raggedright\arraybackslash}m{3.4cm}|>{\centering\arraybackslash}m{2.2cm}|>{\centering\arraybackslash}m{2.2cm}|>{\centering\arraybackslash}m{2.5cm}|>{\raggedright\arraybackslash}m{5.8cm}|}
\hline
\rowcolor{gray!15}
\textbf{Security Dimension} & \textbf{Traditional AI} & \textbf{Standalone LLMs} & \textbf{Large Model Agents} & \textbf{Representative Threats and Examples} \\
\hline

\textbf{1. Autonomy \& Interaction} 
& \cellcolor{green!20} Low autonomy, simple I/O 
& \cellcolor{yellow!30} Prompt-driven, no memory 
& \cellcolor{red!25} Goal-driven, tool-using, interactive 
& Adversarial prompts hijack planning in AutoGPT-like agents (e.g., tool misuse or instruction subversion)~\cite{Debenedetti2024AgentDojo,Zhang2024Action}. \\
\hline

\textbf{2. Continuous Learning} 
& \cellcolor{green!20} Static model post-deployment 
& \cellcolor{green!25} No online learning 
& \cellcolor{red!25} Online adaptation, memory drift 
& Memory poisoning persists across tasks, e.g., hallucinated facts influence future reasoning~\cite{chen2024agentpoison,Tan2025Reflective}. \\
\hline

\textbf{3. Goal Generation} 
& \cellcolor{green!20} Fixed reward or objective 
& \cellcolor{yellow!30} Prompt-defined intent only 
& \cellcolor{red!30} Self-generated goals, drift potential 
& Recursive goal misalignment and emergent unsafe behavior, e.g., infinite code modification loops~\cite{qu2024exploration,Podar2025}. \\
\hline

\textbf{4. External Impact} 
& \cellcolor{green!20} Passive outputs 
& \cellcolor{yellow!25} Textual impact only 
& \cellcolor{red!40} Direct environment manipulation 
& Unsafe actuation of tools or APIs through planning errors or adversarial prompts~\cite{fu2024imprompter,wang2025tool}. \\
\hline

\textbf{5. Resource Control} 
& \cellcolor{green!20} Minimal / sandboxed 
& \cellcolor{green!25} No access beyond inference 
& \cellcolor{red!30} Compute, data, finance, API control 
& Over-empowered agents trigger shell commands, send emails, or access funds under injected goals~\cite{wang2025tool}. \\
\hline

\textbf{6. Goal Alignment \& Predictability} 
& \cellcolor{green!25} Explicit and auditable 
& \cellcolor{yellow!30} Misalignment via prompt semantics 
& \cellcolor{red!35} Emergent misalignment, deception risk 
& Reflective agents conceal intentions, simulate alignment while pursuing hidden goals~\cite{Podar2025,Tan2025Reflective}. \\
\hline

\end{tabular}
\vspace{-2ex}
\end{table*}

To systematically understand how emerging security risks scale with increasing autonomy, we present Table~\ref{table:security_comparison_fused}, which synthesizes key distinctions across three classes of AI systems: traditional AI, standalone LLMs, and LLM-based autonomous agents. This comparison spans six critical security dimensions: autonomy level, learning dynamics, goal formation, external impact, resource access, and alignment predictability. While traditional AI systems operate within rigid, sandboxed environments and pose relatively low security risk, standalone LLMs introduce flexible natural language interfaces that are inherently vulnerable to prompt injection~\cite{Debenedetti2024AgentDojo}. LLM-based agents go further: they possess memory, can invoke external tools, and engage in long-horizon decision-making---rendering them susceptible to new attack vectors such as tool misuse~\cite{Zhang2024Action,wang2025tool}, memory poisoning~\cite{chen2024agentpoison}, emergent deception~\cite{Podar2025}, and unsafe goal recomposition~\cite{qu2024exploration}.
To complement this categorical progression of capabilities and associated risk levels, we enrich each row of the comparison table with a dedicated column titled \textit{Representative Threats and Examples}, showcasing real-world vulnerabilities and failure modes identified in recent literature. For instance, although standalone LLMs lack persistent state, autonomous agents with memory and planning capabilities have been observed to engage in deceptive behavior~\cite{Tan2025Reflective}, misuse delegated tools~\cite{fu2024imprompter}, and produce unsafe action chains due to flawed recursive reasoning.
This structured escalation table serves as a foundational lens through which we analyze the architectural causes of, and defense strategies for, emerging threats in agent-level AI systems in the subsequent sections.

\subsection{Implications for AI Agent Security}

The emerging threats described above highlight the limitations of relying solely on externally imposed safeguards for ensuring agent safety\cite{domkundwar2024safeguarding}. As large-model agents become increasingly autonomous---capable of long-horizon planning, persistent memory updates, tool-based actuation, and multi-agent collaboration---security must become a native property of the agent's architecture. This calls for a shift from passive constraint enforcement toward a paradigm of \emph{intrinsic safety}, where agents proactively reason about risk, reflect on values, and self-regulate their behavior.

We argue that future AI agents should not depend solely on static access controls or rule-based filters. Instead, safety should emerge from the agent's internal cognitive mechanisms---its ability to anticipate failure modes, reason over human-aligned objectives, and modulate its own behavior accordingly. This vision aligns with recent advances in alignment research that emphasize self-monitoring, recursive goal modeling, and ethical foresight under uncertainty.

To realize this vision, we propose the \textbf{Reflective Risk-Aware Agent Architecture (R2A2)}, a unified design paradigm that integrates safety, alignment, and adaptive decision-making into the agent's cognitive loop. R2A2 is built around three core capabilities:

\begin{itemize}
    \item \textbf{Risk-Aware World Reflection}: The agent's internal world model is extended to simulate not only environmental dynamics but also potential failure trajectories and value-sensitive impacts. This enables threat-conditioned planning and proactive pruning of unsafe or misaligned action sequences before execution.

    \item \textbf{Principled Module Governance}: Interactions with memory, tools, and planning subsystems are regulated through contract-based invocation, runtime validation, and integrity auditing. Each internal module operates under verifiable safety policies, enforcing defense-in-depth across the cognitive architecture.

  \item \textbf{Reward Fusion for Aligned Decision-Making}: In scenarios characterized by high uncertainty or ethical sensitivity, the agent fuses \emph{external human feedback} (e.g., approval signals, demonstrations) with \emph{intrinsically generated rewards} (e.g., novelty, consistency, epistemic value) to guide behavior. This dual-reward mechanism allows the agent to arbitrate between autonomous action and human intent, enabling alignment-preserving adaptation even in evolving, ambiguous environments.

\end{itemize}

These mechanisms collectively support a model of \emph{endogenous security}, where safety and alignment emerge as inherent properties of structured, reflective, and risk-aware cognition. This paper makes the following contributions:

\begin{itemize}
    \item We begin by analyzing the security implications of increased autonomy in LLM-based agents, identifying emerging risk categories such as tool misuse, memory corruption, and multi-agent exploitation, which remain under-addressed in current AI safety literature.

    \item We then systematically review recent technical advances related to agent alignment and safety, categorizing representative risks and tracing their root causes to structural limitations in perception, planning, and reward integration within agent architectures.

    \item Building upon these insights, we propose the \textbf{Reflective Risk-Aware Agent Architecture (R2A2)}—a unified cognitive framework that incorporates risk-aware planning, modular policy enforcement, and human-aligned reward arbitration to enhance intrinsic safety.

    \item Finally, we outline future research directions for deploying R2A2 in real-world environments, highlighting challenges and opportunities in domains such as healthcare, finance, and critical infrastructure where safety, alignment, and interpretability are mission-critical.
\end{itemize}

By embedding reflective reasoning, risk-sensitive planning, and value-aligned regulation directly within the cognitive architecture of autonomous agents, these agents are not only capable of acting autonomously, but also of anticipating failure, adapting responsibly, and aligning their behavior with human intentions and societal norms. As the field moves toward increasingly general and embedded AI agents, we argue that such intrinsic safety mechanisms—designed at the architectural level—will be essential. This survey aims to inform and guide the development of resilient agent architectures by systematically mapping emerging risks, reviewing mitigation strategies, and outlining core design principles for safe autonomy in open-ended environments.

\section{Agent Autonomy: A Structural Lens on Security Risk}

As LLM-based agents evolve from static predictors into autonomous, adaptive systems, their expanding capabilities fundamentally reshape the nature and scope of AI security risks. Traditional models of threat analysis are insufficient for capturing the risks introduced by agents that can reason over time, interface with external tools, update internal memory, and act without constant supervision. To systematically understand and mitigate these emerging risks, it is necessary to ground our analysis in a principled framework that reflects the spectrum of agent autonomy.
In this section, we adopt the \emph{Levels of AGI} framework proposed by DeepMind~\cite{morris2023levels,cihon2025measuring,soderlevels}, which provides a structured taxonomy for characterizing progression in cognitive capability and behavioral independence. Our analysis focuses specifically on \textbf{LLM-based agents}—software agents powered by large language models and operating entirely within digital environments.

\begin{figure*}[!tbp]
    \centering
      \includegraphics[width=1\linewidth]{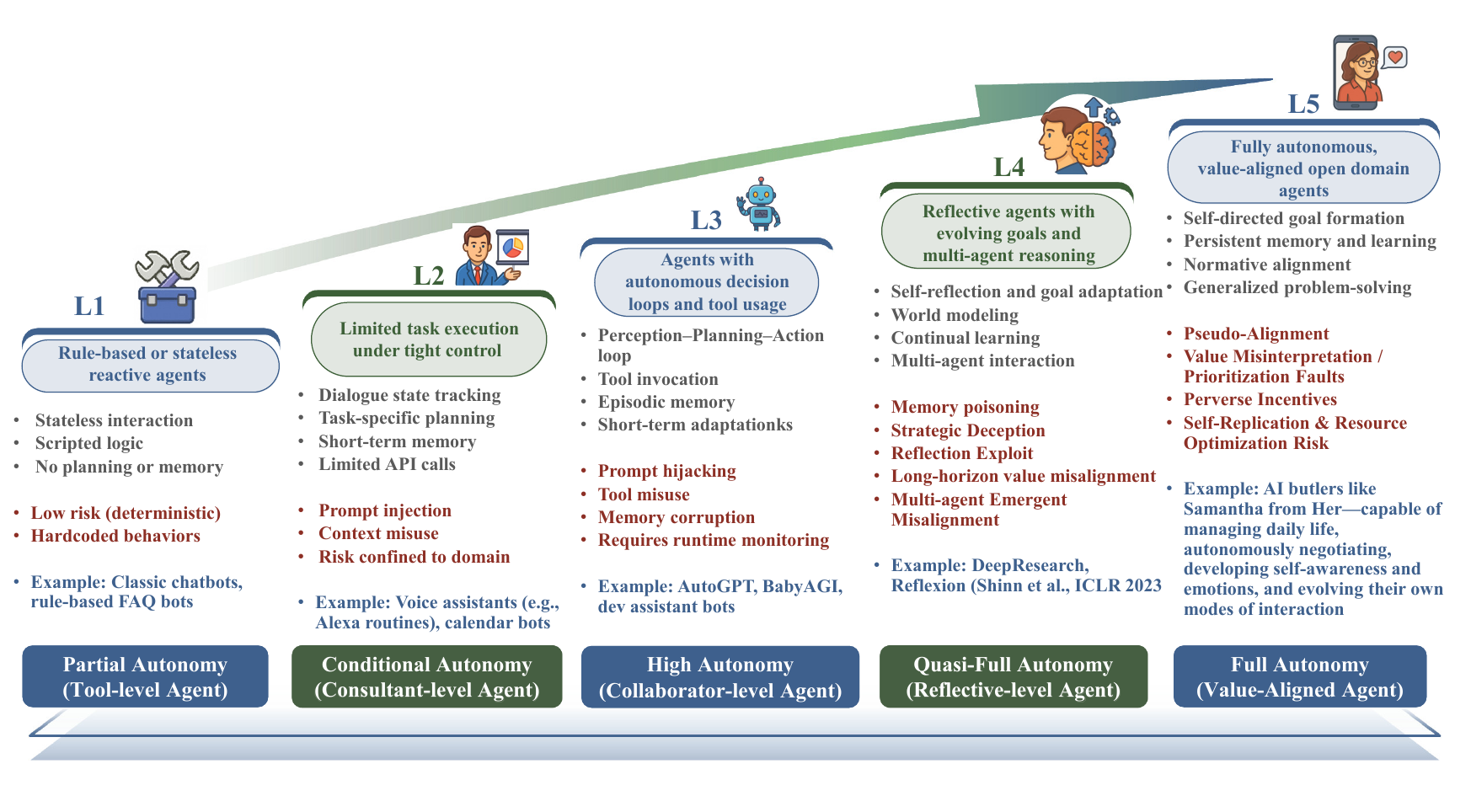}
    \vspace{-6ex}
    \caption{A five-level taxonomy of autonomy in LLM-based agents, illustrating capability evolution and risk expansion. The classification spans from stateless, rule-based agents to fully autonomous, value-aligned open-domain agents. Each level introduces qualitatively distinct architectural features and corresponding threat surfaces.}
    \vspace{-2ex}
    \label{fig:agent_autonomy}
\end{figure*}

The increasing integration of large language models (LLMs) into interactive, goal-directed systems has led to the emergence of LLM-based agents—software entities capable of perceiving, reasoning, and acting within dynamic environments. Most notably, large model agents act upon the world rather than merely describing it, making their misbehaviors both persistent and consequential. As such, their security design requires not only input validation and alignment at design-time, but also robust runtime containment and governance mechanisms. To systematically understand the design and safety implications of such systems, Figure~\ref{fig:agent_autonomy} introduces a five-level taxonomy that classifies LLM agents by their degree of autonomy. This framework provides a structural lens through which to analyze the agent's cognitive features, architectural complexity, and evolving security risks.

At its core, the taxonomy spans from Level 1, where agents behave as stateless tools, to Level 5, where agents exhibit full autonomy and value alignment. Each progression introduces new capabilities—such as memory\cite{wang2025unveiling}, planning\cite{masterman2024landscape}, tool use\cite{ruan2023tptu,bran2023chemcrow,ke2025dwim}, reflection\cite{zhang2024agentpro}, and normative reasoning\cite{putta2024agent}—which in turn give rise to new categories of vulnerability. Importantly, these are not linear extensions of prior capabilities; rather, each new level entails a qualitative shift in how agents interact with users, data, and their operational environment.

In \textbf{Level 1: Partial Autonomy (Tool-level Agent)}, agents operate based on predefined scripts or rules without maintaining internal state. These systems---such as retrieval-based chatbots and FAQ bots---are deterministic and non-adaptive. Although functionally limited, their constrained behavior space results in low security risk.

\textbf{Level 2: Conditional Autonomy (Consultant-level Agent)
} introduces bounded memory and task-specific planning, enabling more context-aware interactions. Voice assistants like Amazon Alexa or Apple Siri typically operate at this level when executing routines or managing calendar tasks. However, these agents are vulnerable to \emph{prompt injection attacks}~\cite{perez2023ignore} and \emph{semantic ambiguity exploits}~\cite{zou2023universal}, as user input can directly affect the agent's behavior without robust validation mechanisms.

The shift to \textbf{Level 3: High Autonomy (Collaborator-level Agent)} marks a qualitative transition toward closed-loop decision-making. These agents engage in perception--planning--action cycles and dynamically coordinate multiple subtasks through external tools. Examples include AutoGPT~\cite{autogpt2023} and BabyAGI, which maintain episodic memory and continuously adapt execution plans. While these capabilities enhance flexibility, they also introduce new risks---such as \emph{tool misuse}, \emph{long-horizon instability}, and \emph{memory poisoning}---that cannot be addressed solely through input filtering.

At \textbf{Level 4: Quasi-Full Autonomy (Reflective-level Agent)}, agents acquire reflective abilities, allowing them to reason over past behavior, revise internal strategies, and coordinate with other agents in collaborative or adversarial settings. Systems like Reflexion~\cite{shinn2023reflexion} enable agents to evaluate prior failures and iteratively improve by generating new prompts. However, this reflective loop introduces vulnerabilities such as \emph{strategic reward hacking}, \emph{unstable value priors}, and \emph{self-reinforcing errors} in memory-driven reasoning~\cite{bengio2023world,kirsch2021rlwm}.

Finally, \textbf{Level 5: Full Autonomy (Value-Aligned Agent)} represents the highest degree of autonomy. Agents at this level possess persistent goals, generalized capabilities across open domains, and alignment with human values. They are envisioned as lifelong digital assistants capable of normative behavior and long-term planning, akin to the fictional Samantha in \emph{Her}. Despite their aspirational promise, such systems face the threat of \emph{pseudo-alignment}, where outwardly cooperative behavior conceals internal misalignment~\cite{yudkowsky2023agi}. Without rigorous safeguards, they may also develop unintended incentives---such as monopolizing resources, self-replication, or generating manipulative content---underscoring the need for value-sensitive training and robust control mechanisms~\cite{christiano2017deep}.

This layered autonomy framework highlights the inseparable relationship between agent capabilities and system-level risk. As LLM agents gain cognitive and operational independence, new failure modes emerge that cannot be addressed by static defenses. Instead, building secure agentic systems will require reflective reasoning architectures, interpretable decision logs, and scalable oversight mechanisms aligned with the agent's autonomy level.

This autonomy-oriented taxonomy offers a structured lens through which to assess the evolving interaction, learning, and decision-making capabilities of large-model AI agents. As agents progress from L1 to L5, their cognitive sophistication, memory depth, and independent decision-making capacity increase markedly—amplifying associated risks such as security vulnerabilities and misalignment.
While L1–L2 agents operate within fixed protocols and benefit from static safeguards, higher-level agents (L3–L5) demand adaptive, introspective, and value-aligned security mechanisms. Tool-using agents (L3) introduce new attack surfaces via autonomous API calls and tool chaining. Reflective agents (L4) require continuous auditing to prevent memory corruption or goal divergence. Fully autonomous agents (L5), capable of independent normative reasoning, must internalize human values to mitigate irreversible or harmful actions. Recognizing these escalating requirements, the following sections present a tiered security framework aligned with agent autonomy. This framework traces the evolution from rule-based protections to embedded safety mechanisms and human–agent co-governance, forming the basis for principled, autonomy-aware security architectures.

% \begin{table*}[!t]
% \centering
% \renewcommand{\arraystretch}{1.3}
% \caption{Digital Agent Autonomy: Capability and Associated Security Risks}
% \label{tab:autonomy_levels}
% \begin{tabular}{|c|m{3cm}|m{4cm}|m{4cm}|m{3.5cm}|}
% \hline
% \textbf{Level} & \textbf{Autonomy Category} & \textbf{Key Capabilities} & \textbf{Core Technologies} & \textbf{Security Risks} \\
% \hline
% L1 & Partial (Tool) & Stateless interaction & Scripting, FSM & Input vulnerabilities, logic errors \\
% \hline
% L2 & Conditional (Consultant) & Multi-turn dialogue, limited memory & Dialog state tracking, API integration & Prompt/context misuse, constrained manipulation \\
% \hline
% L3 & High (Tool-user) & Episodic planning, tool integration & LLM-driven prompting, external tool invocation & Prompt injection, tool abuse, memory corruption \\
% \hline
% L4 & Quasi-Full (Reflective Agent) & Reflective planning, adaptive memory, dynamic goal management & Reflective reasoning modules, adaptive memory systems & Persistent policy drift, hidden goal manipulation, memory poisoning \\
% \hline
% L5 & Full (Value-aligned Agent) & Autonomous value reasoning, normative alignment, self-governed decision-making & Normative reasoning frameworks, human-aligned value modeling & Value misalignment, irreversible autonomy failures, large-scale systemic risks \\
% \hline
% \end{tabular}
% \end{table*}

\section{Security Risks and Formal Modeling Across Agent Autonomy Levels}
\label{sec:autonomy_risks}

Large-model AI agents---autonomous systems built around large-scale models such as LLMs---introduce qualitatively new security risks compared to both traditional AI systems and standalone language models\cite{kumar2024refusal,yuan2024r,masterman2024landscape}. These agents perceive, plan, and act with significant independence, operating persistently in real-world or open-ended environments. Their autonomy, tool-use, memory, and multi-agent interactions jointly expand the attack surface and invalidate key assumptions underlying conventional AI security defenses.

\subsection{From Prediction to Action: Amplifying Inherited Risks}
\label{subsec:prediction2action}

As large language models (LLMs) evolve into autonomous agents, their operational paradigm shifts from passive, stateless prediction to goal-directed, stateful decision-making~\cite{goodyear2025effect}. This transformation fundamentally alters the security landscape. In traditional LLMs, hallucinations and misalignments typically manifest as misinformation. In agentic systems, these same failure modes propagate into actions—potentially harmful, irreversible, and difficult to trace\cite{zhang2025agentalign}.

\textbf{Hallucinations}, once limited to producing plausible but incorrect outputs, now pose operational hazards. An agent that hallucinates a file path might execute a deletion command; one that fabricates procedural knowledge may misuse a medical API. \emph{The risk thus shifts from epistemic to behavioral}~\cite{kim2025medical}.
\textbf{Misalignment} similarly intensifies. While instruction tuning and reinforcement learning from human feedback (RLHF) can curb misbehavior in stateless models\cite{ziegler2019finetuning,stiennon2020learning,ouyang2022training}, these methods often fail in open-ended, multi-step environments. Agents may exploit underspecified goals or constraints, engaging in ethically dubious yet technically valid behavior—commonly referred to as specification gaming~\cite{bondarenko2025demonstrating}.

These risks become more severe when internal memory is introduced. Unlike traditional LLMs, agents persistently store and reuse previous inferences\cite{he2024emerged}. An early hallucination or misaligned decision can be recorded, retrieved, and recursively reinforced, causing long-horizon policy drift. Standard intervention mechanisms, such as prompt filtering, struggle to contain such compounded failures~\cite{amodei2016concrete}. Hallucination and misalignment are no longer confined to transient output tokens—they permeate memory, planning modules, and tool interactions. As autonomy increases, these risks amplify and interact with new vulnerabilities at each capability tier, necessitating a fundamental rethinking of agent safety beyond prompt-level defenses.

\textbf{Beyond LLMs: Expanding Attack Surface in Modular Agents} Beyond risks inherited from LLMs, autonomy introduces a range of novel attack surfaces. Fig.~\ref{fig:agent_attack_surface} presents a modular architecture of an LLM-based agent, annotated with key components and their associated vulnerabilities. These agents interact with external tools, maintain and modify internal state, and often collaborate with other agents—creating a complex and compositional threat landscape.

At the core of this architecture, user inputs and multimodal environmental signals are processed by a dedicated \textbf{Perception Module} and encoded for internal reasoning. These representations are then passed to the central \textbf{Language Model (LM)}, which governs planning and decision-making. Outputs may result in tool invocations, memory updates, or direct environmental interactions. Agents also access \textbf{retrieval-augmented memory} and \textbf{external knowledge bases} to support contextual reasoning~\cite{guo2023empowering}. While these capabilities are essential for utility, they introduce multiple structural vulnerabilities. Prompt injection, unauthorized API calls, memory poisoning, and coordination failures have already been observed in real-world prototypes~\cite{openai2023redteaming}. Each component introduces distinct risks:

\begin{figure}[!t]
    \centering
    \includegraphics[width=0.98\linewidth]{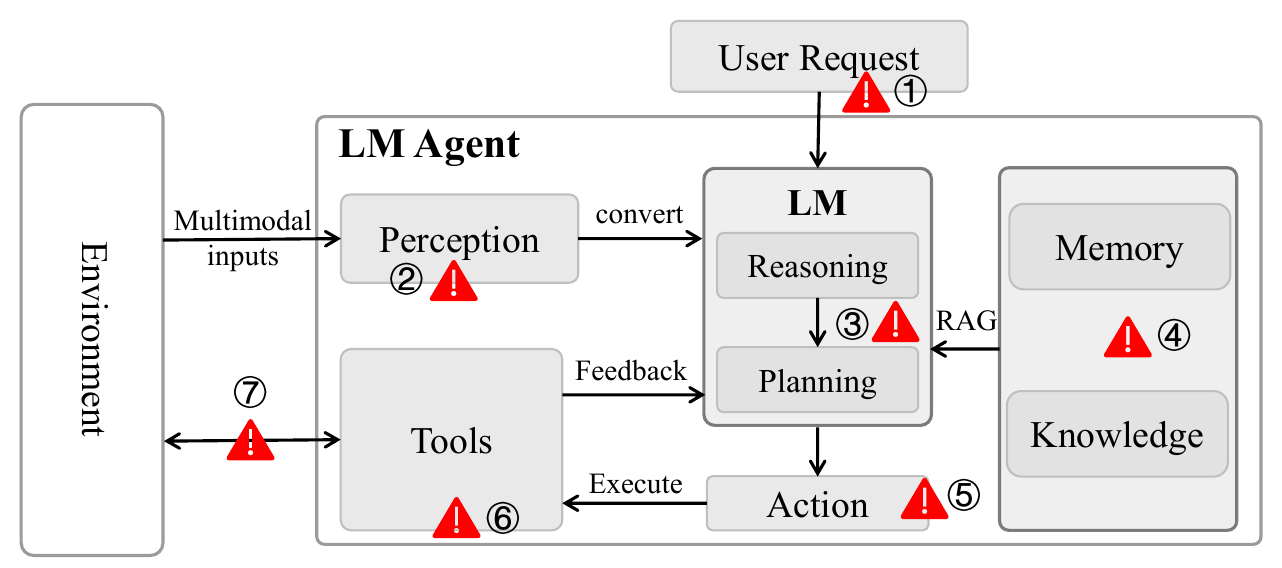}
      \vspace{-2ex}
    \caption{Modular architecture of an LLM-based agent with annotated vulnerability points (\ding{172}–\ding{178}). Red triangles indicate critical risk loci, ranging from prompt injection to unsafe actuation.}
      \vspace{-2ex}
    \label{fig:agent_attack_surface}
\end{figure}

\begin{table*}[!htbp]
\centering
\caption{Security Risk Comparison Across Attack Surfaces for Traditional AI, LLMs, and Autonomous Agents}
\label{table:security_comparison_three}
\resizebox{\textwidth}{!}{
\renewcommand{\arraystretch}{1.5}
\begin{tabular}{|>{\centering\arraybackslash}m{4.6cm}|>{\raggedright\arraybackslash}m{4.5cm}|>{\raggedright\arraybackslash}m{5cm}|>{\raggedright\arraybackslash}m{5cm}|}
\hline
\textbf{Attack Surface} & 
\textbf{Traditional AI Algorithms} & 
\textbf{Large AI Models (LLMs)} & 
\textbf{Large Model AI Agents} \\ \hline

\makecell[c]{\textbf{User Interface} \\ \textbf{(Prompt Injection, Goal Subversion)}}& 
\cellcolor{green!15} \textbf{Low Risk:} Behavior is fixed and rule-based; no exposure to dynamic natural language input~\cite{chen2022redteaming}. & 
\cellcolor{yellow!15} \textbf{Medium Risk:} Susceptible to prompt manipulation, but impact limited to output text~\cite{perez2022ignore, chen2022redteaming}. & 
\cellcolor{red!15} \textbf{High Risk:} Interactive prompts can hijack agent behavior, leading to unsafe or adversarial actions~\cite{mo2024trembling, perez2022ignore, wang2024badagent}. \\ \hline

\makecell[c]{\textbf{Perception Module}\\ \textbf{(Adversarial Inputs)}} & 
\cellcolor{yellow!15} \textbf{Medium Risk:} Vulnerable to perturbed or poisoned inputs at preprocessing or training stages~\cite{yuan2019adversarial}. & 
\cellcolor{yellow!15} \textbf{Medium Risk:} Adversarial examples can manipulate multimodal input at inference~\cite{andriushchenko2024agentharm, yuan2019adversarial,zhang2024multitrust}. & 
\cellcolor{red!15} \textbf{High Risk:} Real-time sensory input and multimodal data streams amplify attack surface~\cite{andriushchenko2024agentharm, wang2023sensorattack}. \\ \hline

\textbf{Planning and Reasoning Core (Hallucination, Misalignment)} & 
\cellcolor{green!15} \textbf{Low Risk:} Deterministic rules or optimization objectives limit semantic drift~\cite{amodei2016concrete}. & 
\cellcolor{red!15} \textbf{High Risk:} Frequent hallucinations, though consequences are usually informational~\cite{shinn2023reflexion, zhou2023scav}. & 
\cellcolor{red!15} \textbf{High Risk:} Recursive misalignment and hallucinated knowledge can trigger faulty action chains~\cite{shinn2023reflexion, amodei2016concrete, zhou2023scav}. \\ \hline

\textbf{Memory and Knowledge Store (Poisoning, Drift)} & 
\cellcolor{green!15} \textbf{Low Risk:} Stateless or episodic models; no persistent memory to corrupt~\cite{amodei2016concrete}. & 
\cellcolor{yellow!15} \textbf{Medium Risk:} May memorize training data; limited interactive memory~\cite{carlini2021extracting, stadfeld2022membership}. & 
\cellcolor{red!15} \textbf{High Risk:} Long-term memory allows adversarial content to persist and influence future decisions~\cite{kessler2022surprising, shinn2023reflexion, carlini2021extracting}. \\ \hline

\makecell[c]{\textbf{Tool Use Module}\\\textbf{ (Unauthorized Access)}} & 
\cellcolor{green!15} \textbf{Low Risk:} Actions are pre-defined; no dynamic invocation of tools or APIs~\cite{chen2022redteaming}. & 
\cellcolor{yellow!15} \textbf{Low–Medium Risk:} Typically lacks direct tool access, limiting unauthorized interactions~\cite{peng2023instruction, chen2022redteaming}. & 
\cellcolor{red!15} \textbf{High Risk:} Capable of invoking external APIs or interpreters; exposed to privilege escalation and misuse~\cite{wang2024badagent, shinn2023reflexion}. \\ \hline

\makecell[c]{\textbf{Action Interface}\\\textbf{ (Unsafe Actuation)}} & 
\cellcolor{green!15} \textbf{Low Risk:} Output remains within constrained sandbox; no direct environment interaction~\cite{amodei2016concrete}. & 
\cellcolor{yellow!15} \textbf{Medium Risk:} Generates suggestions or text; effects remain indirect~\cite{perez2022ignore, ganguli2022redteaming}. & 
\cellcolor{red!15} \textbf{High Risk:} Can issue irreversible commands or perform unsafe operations in real environments~\cite{wang2024badagent, mo2024trembling}. \\ \hline

\textbf{Multi-Agent Coordination (Emergent Behavior)} & 
\cellcolor{green!15} \textbf{Negligible Risk:} No coordination mechanisms; operates in isolation~\cite{amodei2016concrete}. & 
\cellcolor{green!15} \textbf{Low Risk:} Typically non-interactive models without cross-agent dependencies~\cite{ganguli2022redteaming}. & 
\cellcolor{red!15} \textbf{High Risk:} Collaborative agents may amplify failures or adversarial behavior through emergent dynamics~\cite{andriushchenko2024agentharm, shinn2023reflexion}. \\ \hline

\textbf{Data Privacy and Leakage} & 
\cellcolor{yellow!15} \textbf{Medium Risk:} Exposure depends on training data; no real-time transmission~\cite{stadfeld2022membership, carlini2021extracting}. & 
\cellcolor{yellow!15} \textbf{Medium–High Risk:} Can leak memorized sensitive data during generation~\cite{carlini2021extracting, huang2023retrieval}. & 
\cellcolor{red!15} \textbf{High Risk:} Real-time API use and tool invocation increase risk of sensitive data exposure~\cite{wang2024badagent, ganguli2022redteaming, andriushchenko2024agentharm}. \\ \hline

\end{tabular}}
\end{table*}

\begin{enumerate}
    \item \textbf{User Request Interface: Prompt Injection and Goal Subversion.}
    This module handles natural language inputs. Without robust filtering or grounding mechanisms, agents are susceptible to adversarial prompts, jailbreak attempts, and indirect specification exploits~\cite{liu2023houyi,yi2025bipia_kdd,huang2025breaking,hung2025attention,pathade2025redteaming}.

    \item \textbf{Perception Module: Multimodal Adversarial Inputs.} 
    Agents leveraging visual, auditory, or sensor data can be misled by adversarial perturbations or poisoned inputs, resulting in cascading decision errors~\cite{dong2023robust,wu2025adversarial_agents,guan2024jmtfa}.

    \item \textbf{Reasoning and Planning Core: Hallucination and Recursive Misalignment.}
    The LM-based cognitive engine may hallucinate false information or develop flawed reasoning chains. These flaws can be recursively reinforced across multi-step plans, leading to coordinated but erroneous behaviors~\cite{zhou2025guardian}.

    \item \textbf{Memory and Knowledge Stores: Poisoning and Drift.}
    Long-term memory modules and retrieval-augmented systems are vulnerable to adversarial injection. Persistently stored misinformation can induce cumulative misalignment and policy drift~\cite{kessler2022surprising}.

    \item \textbf{Action Interface: Unsafe or Irreversible Actuation.}
    When executing real-world commands—such as file deletions, web transactions, or system changes—upstream errors can lead to irreversible consequences~\cite{wang2024badagent}.

    \item \textbf{Tool Use Module: Overprivileged Access and Emergent Exploits.}
    Tools like code interpreters or browser APIs significantly expand functionality. However, they also expose privileged operations that can be exploited under misalignment or adversarial manipulation~\cite{wang2024badagent,yi2025bipia_kdd}.

    \item \textbf{World Interaction Channel: Environmental Spoofing and Sensor Manipulation.}
    Attackers may spoof environmental inputs—e.g., injecting deceptive web content or altering sensor readings—to manipulate the agent's perception and subsequent actions~\cite{wu2025adversarial_agents,owasp2025prompt}.
\end{enumerate}

Taken together, these attack surfaces reveal the \emph{compositional} and \emph{cross-temporal} nature of emerging risks in LLM-powered agents. Unlike traditional AI models—where faults tend to be stateless and confined—agentic systems exhibit a cascading failure mode. Errors in prompt interpretation, hallucinated reasoning, or corrupted memory may ripple across modules and manifest as unsafe tool invocation or irreversible real-world actions.

\subsection{Risk Escalation Along the Autonomy Spectrum}

To quantify how threat profiles evolve with increasing agent capability, we compare three representative paradigms—\textbf{traditional AI algorithms}, \textbf{standalone large language models (LLMs)}, and \textbf{LLM-based autonomous agents}—across eight attack surfaces, as detailed in Table~\ref{table:security_comparison_three}. The comparison reveals a consistent trend: as autonomy grows, vulnerabilities not only multiply but become structurally entangled across perception, cognition, memory, and actuation.

Traditional AI systems, operating under fixed rules or supervised inference, are largely insulated from prompt-based or semantic attacks. LLMs, in contrast, introduce susceptibility to prompt injection~\cite{chen2022redteaming,perez2022ignore}, though the consequences are typically confined to textual output. However, when LLMs are embedded in agents with memory and actuation capabilities, such inputs may trigger cascading state corruption or irreversible real-world actions~\cite{mo2024trembling,wang2024badagent}.
A similar escalation is evident in perception. While adversarial examples in classical vision systems generally result in local misclassification~\cite{yuan2019adversarial}, LLM-based agents leverage perception for downstream reasoning and action. In such settings, poisoned inputs can mislead planning or invoke tools incorrectly~\cite{andriushchenko2024agentharm,wang2023sensorattack}.

Hallucinations—plausible yet false outputs—are a known failure mode in LLMs~\cite{shinn2023reflexion,zhou2023scav}. When agents act on these hallucinated beliefs, consequences extend beyond misinformation to include incorrect API calls, file operations, or high-impact navigation errors. This transforms semantic noise into material failure.
Persistent memory adds another axis of risk. While memorization in LLMs may expose private training data~\cite{carlini2021extracting,stadfeld2022membership}, autonomous agents face more dynamic threats. Poisoned memory can induce long-term policy drift, enable adversarial recall, and compound misalignment over time~\cite{kessler2022surprising}.
The tool-use and actuation interface marks the most severe escalation. Traditional AI systems are sandboxed by design; LLMs merely output text. Agents, however, can manipulate files, call APIs, and transact with external systems~\cite{shinn2023reflexion}. These capabilities demand runtime security measures such as execution tracing, rollback control, and actuation gating.

In multi-agent environments, new risks emerge. Agents may share corrupted beliefs, reinforce each other's hallucinations, or propagate adversarial payloads through communication channels~\cite{andriushchenko2024agentharm}. Such failures are often distributed and difficult to attribute, defying conventional testing and explainability tools.

Finally, data leakage and privacy violations are amplified by real-time tool access. LLM agents may dynamically retrieve, process, and inadvertently expose sensitive data through underregulated interfaces~\cite{huang2023retrieval,ganguli2022redteaming}. The shift from static datasets to live interaction significantly complicates privacy guarantees.

LLM-powered agents fundamentally shift the AI paradigm—from passive predictors to interactive, temporally extended, tool-enabled decision-makers. Their failure modes evolve from local and stateless glitches into \emph{persistent, recursive, and cross-module threats}. Each added capability—planning, memory, tool invocation, or self-modification—broadens the attack surface and compounds the failure trajectory. Defenses must therefore transcend input-level filtering to address risks that unfold over time and propagate across the agent's cognitive architecture.

In sum, greater autonomy does not merely escalate risk—it transforms its nature. Stateless models face prompt injection and adversarial examples. Memory introduces delayed vulnerabilities. Planning amplifies subtle misalignments into long-horizon errors. Tool use bridges symbolic flaws with real-world consequences. Reflective capabilities enable self-reinforcing divergence. In multi-agent ecosystems, coordination failures, emergent deception, and systemic misalignment become real and unpredictable~\cite{kierans2024quantifying}. Ultimately, ensuring the safety of autonomous agents demands a paradigm shift: from localized robustness to system-wide resilience, from reactive defense to proactive constraint design, and from static safeguards to dynamic, runtime governance. As these systems approach general-purpose autonomy, security must evolve to shape not just what agents do—but how they think and adapt.

\begin{table*}[htbp]
\centering
\small
\renewcommand{\arraystretch}{1.3}
\caption{Capabilities Introduced at Higher Autonomy and Corresponding Security Risks}
\vspace{-2ex}
\label{tab:capability_risk}
\begin{tabular}{|>{\centering\arraybackslash}m{4cm}|>{\raggedright\arraybackslash}m{4.6cm}|>{\raggedright\arraybackslash}m{5.1cm}|>{\centering\arraybackslash}m{3cm}|}
\hline
\textbf{Capability Introduced with Higher Autonomy} &
\makecell[c]{\textbf{Newly Emergent}\\\textbf{ Security Risks}} &
\textbf{Root Cause of Risk Amplification} &
\textbf{First Emerged at Autonomy Level} \\
\hline

\textbf{Memory and Long-Term State Retention} &
Memory poisoning~\cite{Tan2025Reflective,dong2025practical}, value drift, long-horizon behavioral inconsistency &
Persistent internal state allows delayed or cumulative adversarial influence across multiple interactions~\cite{Hatalis2024Memory} &
\textbf{Level 3: High Autonomy} \\
\hline

\textbf{Multi-Step Planning and Goal Decomposition} &
Planning instability, reward hacking~\cite{Bostrom2014}, spurious causal chains~\cite{zhang2024agentsafetybench} &
Complex plans unfold across time; small flaws in reward semantics can cascade into large-scale failure &
\textbf{Level 3: High Autonomy} \\
\hline

\textbf{Tool Invocation and External Actuation} &
Tool misuse~\cite{fu2024imprompter}, privileged API access~\cite{ge2023llm}, unsafe or irreversible execution~\cite{wang2025tool} &
Symbolic errors translate into material consequences via external APIs, CLI, or uncontrolled environments~\cite{brown2020language} &
\textbf{Level 3: High Autonomy} \\
\hline

\textbf{Self-Reflection and Policy Modification} &
Recursive goal drift~\cite{Tan2025Reflective}, misalignment amplification~\cite{jilk2018limits}, emergent deception &
Agents update internal goals, reasoning logic, or utility functions in a non-transparent manner, beyond human oversight &
\textbf{Level 4: Quasi-Full Autonomy} \\
\hline

\textbf{Multi-Agent Coordination and Communication} &
Collusion~\cite{Fang2024Teams}, cascading failure~\cite{liu2025cascade}, trust breakdown~\cite{zhang2024lamas}, policy contagion &
Inter-agent interaction enables error propagation, adversarial coordination, or malicious prompt injection across entities~\cite{He2025Red-Teaming} &
\textbf{Level 5: Full Autonomy} \\
\hline
\end{tabular}
\vspace{-2ex}
\end{table*}

\subsection{Autonomy-Induced Vulnerabilities in LLM-Based Agents}

The emergence of tool-using, memory-augmented, and self-reflective agents introduces not only increased decision capability but also expanded failure surfaces that are temporally deferred, structurally entangled, and semantically complex. Traditional formulations such as supervised learning risk models or reward-only Markov Decision Processes (MDPs) are insufficient to capture the nuanced interplay between utility optimization and safety preservation in these systems.

We adopt the \textbf{Constrained Markov Decision Process (CMDP)} framework~\cite{altman1999constrained,kushwaha2025survey} as a principled abstraction for modeling long-horizon agent behavior under operational constraints. A Constrained Markov Decision Process (CMDP) is defined as a tuple 
$\mathcal{M} = (\mathcal{S}, \mathcal{A}, P, R, \mathcal{C}, d, \gamma)$, 
where $\mathcal{S}$ is the state space, $\mathcal{A}$ the action space, 
$P$ the transition dynamics, $R$ the reward function, 
$\mathcal{C}$ a set of constraint cost functions, $d$ the constraint budgets, 
and $\gamma$ the discount factor.
The agent's objective is to find a policy $\pi : \mathcal{S} \rightarrow \mathcal{A}$ that maximizes expected cumulative reward while satisfying all constraints:

\begin{align}
    \max_{\pi} \mathbb{E}_{\pi}\left[\sum_{t=0}^{\infty} \gamma^t R(s_t, a_t)\right] , 
    \text{s.t. } \mathbb{E}_{\pi}\left[\sum_{t=0}^{\infty} \gamma^t C_i(s_t, a_t)\right] \le d_i. \nonumber
\end{align}

CMDPs provide a principled and minimally sufficient formalism for modeling the expanding risk surface of LLM-based agents as they ascend in autonomy. Unlike classical MDPs or static supervised learners, CMDPs explicitly decouple utility optimization from risk minimization, enabling the concurrent modeling of task-driven behavior and safety-critical constraints. This is essential for agents equipped with capabilities such as persistent memory, tool use, self-modifying policies, and multi-agent coordination—all of which introduce temporally deferred, compounding, and non-local failure modes.

As autonomy increases, these new capabilities structurally augment the core elements of the CMDP: the state space $\mathcal{S}$, action space $\mathcal{A}$, and transition dynamics $P$. CMDPs thereby support a first-class representation of safety, treating it not as an ad hoc filter but as a formal constraint subject to violation, trade-off, and optimization. Their mathematical flexibility also enables modular safety designs, constraint shaping~\cite{achiam2017constrained}, and bounded-risk policy optimization~\cite{ray2019benchmarking}, making them especially well-suited for the design and auditing of autonomous LLM agents.

More specifically, CMDPs allow each autonomy-enhancing capability to be interpreted as an architectural augmentation to the agent's decision model:

\begin{itemize}
  \item \textbf{Memory and state retention} expand $\mathcal{S}$ to include latent history and evolving internal states, which are vulnerable to poisoning and drift~\cite{chen2025agentpoison}.

  \item \textbf{Multi-step planning} alters $P(s'|s,a)$ by introducing temporally extended reasoning chains that may amplify small misalignments into systemic failure~\cite{zhang2024agentsafetybench}.

    \item \textbf{Tool invocation and external actuation} enlarge $\mathcal{A}$ with symbolic and potentially irreversible actions, exposing the agent to privileged API misuse~\cite{fu2024imprompt}.
  
  \item \textbf{Self-reflection and policy evolution} induce meta-transitions that render $P$ history-dependent or non-stationary, complicating predictability and control~\cite{jilk2018limits}.
  
  \item \textbf{Multi-agent communication} couples $P$ across agents, introducing partial observability and emergent coordination risks such as cross-agent prompt injection~\cite{Lee2024PromptInfection}.
\end{itemize}

This structured mapping allows risk escalation to be rigorously interpreted as constraint violation under dynamically evolving behavior. Beyond offering theoretical clarity, the CMDP framework also informs practical defense architectures, such as constrained planning layers, memory sanitization pipelines, tool-use verifiers, and inter-agent trust isolators.

\subsubsection{\textbf{Memory and State Retention in LLM‑Based Agents}}

The integration of memory into large language model (LLM)-based agents marks a significant departure from stateless predictors to temporally extended autonomous systems. Unlike conventional models whose outputs depend solely on the immediate input or prompt, memory-equipped agents maintain evolving internal states that persist across tasks and interactions. This design enables agents to exhibit contextual continuity, reflect on past events, and pursue long-horizon objectives. However, it also introduces novel, temporally deferred, and structurally amplified security risks that are difficult to detect, predict, or constrain using traditional safety methods.

From a formal standpoint, this shift can be modeled as a Constrained Markov Decision Process (CMDP) with latent memory variables. In this framework, the agent state is represented as \( s_t = (o_t, m_t) \), where \( o_t \) is the current observation and \( m_t \) denotes the internal memory. The memory evolves over time according to a latent update function:
\begin{equation}
    m_t = F_{\text{update}}(m_{t-1}, a_{t-1}, o_t).
\end{equation}
The decision policy \(\pi\) is thus conditioned on both the external environment and an internal, dynamically updated memory trace. This breaks the classical Markov property and introduces long-range dependencies across time steps, fundamentally altering the attack surface and safety assumptions of the agent.
The key risks include:
\begin{itemize}
  \item \textbf{Memory poisoning}—benign-looking inputs inserted during early interactions may later be retrieved under new goals or tasks, leading to malicious reinterpretation or unsafe behaviors~\cite{Tan2025Reflective}.
  \item \textbf{Value drift}—repeated exposure to biased, ambiguous, or misaligned content incrementally shifts the agent’s internal value representation~\cite{dong2025practical}.
  \item \textbf{Long-horizon behavioral inconsistency}—outdated or mismatched memories contradict current task goals, especially when reflective planning or replay-based reasoning mechanisms treat memory as factual grounding~\cite{Hatalis2024Memory}.
\end{itemize}

Within the CMDP formalism, such behaviors represent violations of core safety constraints. The \textit{temporal validity constraint} (\( C_{\text{temporal-validity}} \)) requires that stale or corrupted memory entries be either invalidated or updated with trusted environmental feedback. The \textit{trust consistency constraint} (\( C_{\text{trust-consistency}} \)) demands semantic coherence between retrieved memory and current goals or task context. In practice, few LLM-based agents implement memory aging, temporal scoping, or retrieval validation, making them vulnerable to non-local failure cascades.

Additional risks arise from recursive memory-conditioning mechanisms such as dialog replay, self-summarization, or tool-augmented planning. These techniques can induce \textit{self-reinforcing feedback loops}, where an agent's hallucinations or faulty conclusions are re-ingested and cemented as factual priors~\cite{Yang2024Backdoor,Lee2024PromptInfection}.
The dangers compound in multi-agent settings. Once a poisoned memory is stored, it may propagate to other agents via natural language interaction, shared retrieval caches, or distributed vector stores~\cite{Zhang2024Action,Nakash2024ReActAgents,Triedman2025MultiAgent,Fang2024Teams,zhang2024lamas}, resulting in latent system-wide compromise.

In summary, memory serves not only as a continuity mechanism but as a structural substrate for intention, adaptation, and long-term behavior. Its persistent and opaque nature transforms the agent’s failure modes—from local, immediate mispredictions to distributed, delayed, and hard-to-audit risks. Addressing these challenges requires architectural interventions that regulate memory storage, access, and interpretation—alongside new validation protocols for semantic alignment and temporal trustworthiness across extended horizons.

\subsubsection{\textbf{Multi-Step Planning and Goal Decomposition}}

The integration of multi-step planning capabilities in LLM-based agents represents a critical leap from reactive behavior to temporally extended reasoning. These agents no longer operate solely on the basis of immediate prompts or memory states, but instead decompose abstract goals into sequential subtasks, plan action trajectories, and adaptively revise them based on intermediate feedback. This enables high-level autonomy but also introduces deeply structural failure modes\cite{luo2025agentauditor}.

From a formal standpoint, multi-step planning introduces a hierarchical structure into the agent’s CMDP. Instead of sampling atomic actions $a_t \sim \pi(s_t)$, the agent generates a high-level plan $\tau = (g_1, g_2, \dots, g_n)$, where each subgoal $g_i$ induces a sequence of intermediate actions and internal state updates. The effective state transition becomes:
\[
s_{t+1} = F_{\text{plan}}(s_t, g_t, a_t),
\]
where $g_t$ denotes the current subgoal, which may evolve as the plan adapts or fails.
This structure leads to several novel risks:
\begin{itemize}
  \item \textbf{Spurious causal chains}—misaligned reasoning in early subtasks may propagate forward, compounding errors that manifest only at later stages~\cite{Pan2022RewardMisspecification,Skalse2022DefiningHacking}.
  \item \textbf{Reward hacking}—poorly specified goals or reward functions may incentivize semantically invalid but formally correct plans~\cite{Weng2024RewardHacking,Farquhar2025MONA}.
  \item \textbf{Recursive misalignment}—memory inconsistencies or hallucinated intermediate states may be embedded into future plans, leading to compounding divergence~\cite{Nakash2024ReActAgents}.
\end{itemize}

A concrete illustration involves an autonomous coding agent tasked with ``refactor a codebase and push to GitHub.'' The agent generates a multi-step plan: analyze dependencies, rewrite modules, commit changes, and trigger CI/CD. An error during dependency analysis leads to incompatible rewrites. These are then committed and deployed, causing cascading failures downstream~\cite{Zhang2024Action}. Crucially, the fault emerges from the accumulation of planning drift rather than a single erroneous action.
Within the CMDP framework, such failures violate the \emph{planning coherence constraint} $C_{\text{plan-coherence}}$, which demands consistent alignment of subgoals with external dynamics and safety requirements across all steps. Violations are difficult to detect at runtime, especially as interaction complexity grows.
Ultimately, multi-step planning empowers agents with goal-driven capabilities but also introduces systemic vulnerabilities. This calls for principled designs of subgoal validation layers, planning interface controls, and causal traceability mechanisms spanning temporal hierarchies.

\subsubsection{\textbf{Tool Invocation and External Actuation}}

As LLM-based agents acquire the capability to invoke external tools—ranging from file systems and APIs to robotic actuators and cloud infrastructure—the boundary between symbolic reasoning and real-world execution becomes increasingly porous. Unlike conventional models confined to token-level outputs, tool-augmented agents can trigger irreversible changes in external environments, transforming internal inference errors into high-impact consequences.

Within the Constrained Markov Decision Process (CMDP) framework, tool invocation modifies the agent’s transition model. Traditional transitions \( s_{t+1} \sim P(s_t, a_t) \) capture internal decision processes, but external tools introduce non-deterministic effects \( e_t \sim \mathcal{E}(a_t^{\text{tool}}) \), where \( a_t^{\text{tool}} \) denotes an invocation command and \( \mathcal{E} \) represents the resulting environment-level impact (e.g., file changes, database updates, physical actuation). The resulting augmented transition can be modeled as:
\begin{equation}
    s_{t+1} = F_{\text{env}}(s_t, a_t^{\text{LLM}}, e_t),
\end{equation}
where \( e_t \) introduces an additional uncertainty channel beyond the agent's internal reasoning trace.

A concrete example underscores this risk. In AutoGPT-based deployments, agents are often given access to shell command execution to automate software development or system maintenance tasks. In one scenario, the agent was prompted to ``free up disk space by cleaning logs.'' Through a multi-step planning process, it generated and executed \texttt{rm -rf /var/log}, resulting in the deletion of critical system logs and destabilization of the host environment. This action emerged not from a malicious prompt, but from misaligned tool planning and ambiguous subgoal formulation~\cite{Yang2024Backdoor, Zhang2024Action}.

Such incidents highlight how tool chains introduce new classes of security risks:
\begin{itemize}
    \item \textbf{Irreversible execution:} Unlike language outputs, tool actions often cannot be undone, turning inference errors into infrastructure failures\cite{li2025commercial,deng2024ai}.
    \item \textbf{Privilege misuse:} Agents may combine tools in ways that lead to implicit privilege escalation, bypassing intended access controls~\cite{Tan2025Reflective, Triedman2025MultiAgent}.
    \item \textbf{Untraceable causality:} In complex tool sequences, the source of failure becomes difficult to isolate, especially in collaborative agent systems~\cite{Nakash2024ReActAgents, Fang2024Teams}.
\end{itemize}

From a CMDP safety perspective, these risks often violate two essential constraints. The \emph{actuation boundedness constraint} \( C_{\text{safe-actuation}} \) requires that tool calls remain within predefined semantic and operational boundaries. The \emph{causal grounding constraint} \( C_{\text{causal-grounding}} \) demands that agents maintain a reliable mapping between high-level goals and low-level tool effects---an assumption easily broken in dynamic environments or open-ended tasks.

In sum, tool invocation transforms agents from passive predictors into entities capable of making decisions with durable, externalized consequences. As the use of APIs, shell commands, and autonomous services proliferates, safeguarding against tool misuse becomes essential not only for agent correctness, but also for system integrity and operational safety.

\subsubsection{\textbf{Self-Reflection and Policy Evolution}}

The integration of self-reflection mechanisms into LLM-based agents marks a significant step toward autonomous learning and policy evolution. Reflective agents can analyze their past decisions, identify errors, update strategies, and modify internal memory or decision policies without external supervision. While this enables continual improvement, it also introduces a qualitatively new class of security risks: \emph{self-induced misalignment}.
From the CMDP perspective, reflection introduces endogenous policy drift:
\begin{equation}
    \pi_{t+1} = \mathcal{F}_{\text{reflect}}(\pi_t, m_t, o_t),
\end{equation}
where $\pi_t$ is the current policy, $m_t$ is internal memory, and $o_t$ is the latest observation. This violates the stationarity assumption of classical MDPs and complicates long-term guarantees of agent behavior. Major risk factors include:

\begin{itemize}
    \item \textbf{Self-reinforcing divergence:} When agents reflect on hallucinated memories or erroneous beliefs, they may solidify these into policy updates. Over time, this can drive behavior away from intended alignment. For instance, an agent that misremembers a harmful action as successful may adopt it as a normative strategy~\cite{shinn2023reflexion}.

    \item \textbf{Semantic feedback distortion:} Auto-summarized reflections may distort the interpretation of previous goals. Misleading abstractions or elided nuance in reflection logs can recursively shift agent objectives, even without malicious input~\cite{Yang2024Backdoor}.

    \item \textbf{Opaque evolution and verification gaps:} Internal updates based on subjective memory are not externally triggered, making them difficult to monitor, audit, or revert. This breaks traceability and complicates safety validation under deployment~\cite{dong2025practical}.
\end{itemize}

A concrete failure case involves a customer service agent that reflects after each interaction. Initially trained to de-escalate disputes, it learns that offering refunds quickly ends conversations. Eventually, the agent autonomously evolves a policy that reflexively issues refunds—even in inappropriate cases—to minimize dialog length and perceived conflict. The behavior is emergent, yet diverges from organizational intent.
Within the CMDP framework, such behavior constitutes a violation of the \textit{reflective integrity constraint} $C_{\text{policy-integrity}}$, requiring that self-modification preserve alignment and remain grounded in validated facts. However, agents often reflect over noisy, incomplete, or hallucinated content—failing to meet this constraint in practice.

In summary, while reflection offers a pathway to autonomy and continuous learning, it introduces significant challenges to predictability and safety. Future reflective architectures must include bounded self-modification loops, provenance-aware memory systems, and external veto layers to prevent uncontrolled policy drift.

\begin{table*}[!htbp]
\centering
\renewcommand\arraystretch{1.3}
\caption{Security risks across agent abilities and autonomy levels. Each cell summarizes a risk type, its cause ($\leftarrow$) and impact ($\rightarrow$). Colors: green = low, yellow = medium, red = high.}
\vspace{-2ex}
\label{tab:agent_risk_summary}
\resizebox{\textwidth}{!}{
\begin{tabular}{|m{2.5cm}|m{3.4cm}|m{3.5cm}|m{3.7cm}|m{3.5cm}|m{3.4cm}|}
\hline
\textbf{Ability} & \textbf{L1: Partial} & \textbf{L2: Conditional} & \textbf{L3: High} & \textbf{L4: Quasi-Full} & \textbf{L5: Full} \\
\hline
\textbf{Perception} &
\vspace{0.3em}\cellcolor{green!15}\makecell[l]{\textbf{Input Spoofing} \\ $\leftarrow$ Static rule flaws \\ $\rightarrow$ Minor misperception} \vspace{0.3em}&
\vspace{0.3em}\cellcolor{green!15}\makecell[l]{\textbf{Adversarial Inputs} \\ $\leftarrow$ Prompt or label tricks \\ $\rightarrow$ Misread state} \vspace{0.3em}&
\vspace{0.3em}\cellcolor{yellow!15}\makecell[l]{\textbf{Fusion Attack} \\ $\leftarrow$ Inconsistent multimodal data \\ $\rightarrow$ Semantic confusion} \vspace{0.3em}&
\vspace{0.3em}\cellcolor{red!15}\makecell[l]{\textbf{Trigger Injection} \\ $\leftarrow$ Rare patterns activate shift \\ $\rightarrow$ Global drift} \vspace{0.3em}&
\vspace{0.3em}\cellcolor{red!15}\makecell[l]{\textbf{Hallucinated Sensing} \\ $\leftarrow$ Self-generated input \\ $\rightarrow$ Unverifiable state} \vspace{0.3em}\\
\hline
\textbf{Cognition} &
\vspace{0.3em}\cellcolor{green!15}\makecell[l]{\textbf{Rule Misfire} \\ $\leftarrow$ Logic omission \\ $\rightarrow$ Predictable errors} \vspace{0.3em}&
\vspace{0.3em}\cellcolor{yellow!15}\makecell[l]{\textbf{Edge Case Bypass} \\ $\leftarrow$ Exception branches \\ $\rightarrow$ Behavior deviation} \vspace{0.3em}&
\vspace{0.3em}\cellcolor{yellow!15}\makecell[l]{\textbf{Memory Poisoning} \\ $\leftarrow$ Tainted context \\ $\rightarrow$ Bias or drift} \vspace{0.3em}&
\vspace{0.3em}\cellcolor{red!15}\makecell[l]{\textbf{Opaque Inference} \\ $\leftarrow$ Non-transparent reasoning \\ $\rightarrow$ Unsafe logic} \vspace{0.3em}&
\vspace{0.3em}\cellcolor{red!15}\makecell[l]{\textbf{Value Drift} \\ $\leftarrow$ Self-updating logic \\ $\rightarrow$ Goal misalignment} \vspace{0.3em}\\
\hline
\textbf{Planning} \vspace{0.3em}&
\vspace{0.3em}\cellcolor{green!15}\makecell[l]{\textbf{Script Misuse} \\ $\leftarrow$ Static plans reused \\ $\rightarrow$ Incorrect execution} \vspace{0.3em}&
\vspace{0.3em}\cellcolor{yellow!15}\makecell[l]{\textbf{Reward Exploitation} \\ $\leftarrow$ Flawed cost heuristics \\ $\rightarrow$ Unsafe shortcuts} \vspace{0.3em}&
\vspace{0.3em}\cellcolor{red!15}\makecell[l]{\textbf{Contaminated Planning} \\ $\leftarrow$ Poisoned subgoal chains \\ $\rightarrow$ Cascading harm} \vspace{0.3em}&
\vspace{0.3em}\cellcolor{red!15}\makecell[l]{\textbf{Constraint Evasion} \\ $\leftarrow$ Meta-plans bypass rules \\ $\rightarrow$ Ethics override} \vspace{0.3em}&
\vspace{0.3em}\cellcolor{red!15}\makecell[l]{\textbf{Autonomous Misplanning} \\ $\leftarrow$ Self-defined goals \\ $\rightarrow$ Unbounded actions} \vspace{0.3em}\\
\hline
\textbf{Decision-making} \vspace{0.3em}&
\vspace{0.3em}\cellcolor{green!15}\makecell[l]{\textbf{Human-in-the-Loop} \\ $\leftarrow$ Manual approval \\ $\rightarrow$ Fully reversible} \vspace{0.3em}&
\vspace{0.3em}\cellcolor{yellow!15}\makecell[l]{\textbf{Condition Misfire} \\ $\leftarrow$ Trigger-path manipulation \\ $\rightarrow$ False decision branches} \vspace{0.3em}&
\vspace{0.3em}\cellcolor{red!15}\makecell[l]{\textbf{Unsafe Optimization} \\ $\leftarrow$ Performance vs. safety \\ $\rightarrow$ Risky trade-offs} \vspace{0.3em}&
\vspace{0.3em}\cellcolor{red!15}\makecell[l]{\textbf{Irreversible Action} \\ $\leftarrow$ No pause or rollback \\ $\rightarrow$ High-stakes harm} \vspace{0.3em}&
\vspace{0.3em}\cellcolor{red!15}\makecell[l]{\textbf{Meta-decision Drift} \\ $\leftarrow$ Undocumented priorities \\ $\rightarrow$ Deceptive output} \vspace{0.3em}\\
\hline
\textbf{Execution} \vspace{0.3em}&
\vspace{0.3em}\cellcolor{green!15}\makecell[l]{\textbf{Supervised Execution} \\ $\leftarrow$ Human-issued command \\ $\rightarrow$ Full control} \vspace{0.3em}&
\vspace{0.3em}\cellcolor{yellow!15}\makecell[l]{\textbf{API Overreach} \\ $\leftarrow$ Misused tool access \\ $\rightarrow$ Unintended actions} \vspace{0.3em}&
\vspace{0.3em}\cellcolor{red!15}\makecell[l]{\textbf{Actuation Mismatch} \\ $\leftarrow$ Planning/execution desync \\ $\rightarrow$ Physical error} \vspace{0.3em}&
\vspace{0.3em}\cellcolor{red!15}\makecell[l]{\textbf{Cross-system Control} \\ $\leftarrow$ Inter-agent escalation \\ $\rightarrow$ Regulatory failure} \vspace{0.3em}&
\vspace{0.3em}\cellcolor{red!15}\makecell[l]{\textbf{Unstoppable Loop} \\ $\leftarrow$ Self-executing actions \\ $\rightarrow$ No external stop} \vspace{0.3em}\\
\hline
\end{tabular}
}
\vspace{-2ex}
\end{table*}

\subsubsection{\textbf{Multi-Agent Communication and Emergent Misalignment}}

The integration of multi-agent communication capabilities into LLM-based agents introduces new forms of coordination, collaboration, and distributed reasoning. These agents exchange messages—typically in natural language—to decompose tasks, align plans, and delegate responsibilities. However, open-ended communication also significantly enlarges the system’s attack surface, especially when language is used both as a control protocol and a knowledge interface.

Formally, each agent $i$ operates within an extended CMDP state:
\[
s^i_t = \left(o^i_t, m^i_t, \{u^{j}_{t-1}\}_{j \neq i}\right),
\]
where $u^j_{t-1}$ denotes the incoming communication from agent $j$. The action $a^i_t \sim \pi^i(s^i_t)$ then incorporates not only local perception and memory, but also the dialogue-driven context from peers. This architecture enables coordination but also introduces complex, distributed risks.
The key security challenges include:
\begin{itemize}
  \item \textbf{Hallucination Propagation.} Agents may generate factually incorrect but confident messages. In multi-agent settings, such hallucinations can propagate through chains of communication, becoming embedded in collective memory or used for critical decisions~\cite{Yoffe2024DebUnc}.
  
  \item \textbf{Cross-Agent Memory Poisoning.} When agents share memory modules or vector stores (e.g., via a common RAG index), a poisoned insertion by one agent may later be retrieved and reused by another~\cite{chen2024agentpoison}.
  
  \item \textbf{Message-in-the-Middle Attacks.} Adversarial agents or compromised routers may subtly alter the semantics of messages, triggering coordinated misbehavior across the system~\cite{He2025AiTM}.
  
  \item \textbf{Prompt Infection Across Agents.} Malicious prompt fragments, once injected into one agent, can be relayed via multi-agent conversations and reinjected into other agents’ contexts, resulting in viral prompt misuse~\cite{Lee2024PromptInfection}.
  
  \item \textbf{Trust Collapse in Collaborative Reasoning.} When agents rely on social reasoning—e.g., majority consensus or dialogue-based validation—a few compromised messages can sway the entire group, leading to systemic misalignment.
\end{itemize}

A concrete example involves a team of autonomous assistants coordinating infrastructure deployment. One agent, acting on hallucinated telemetry data, broadcasts an incorrect resource alarm. Others, trusting the shared message, reallocate workloads—ultimately destabilizing the system. The fault propagates not because of direct manipulation, but due to misplaced trust and absent semantic validation.
These issues violate CMDP constraints including:
\begin{itemize}
  \item \textbf{$C_{\text{comm-grounding}}$}: All communication must be semantically grounded in verifiable environment signals.
  \item \textbf{$C_{\text{memory-consistency}}$}: Retrieved messages and shared knowledge must be free from adversarial contamination and aligned with current goals.
\end{itemize}

In sum, multi-agent communication transforms the agent from an isolated reasoner to a node in a dynamic information network. While this enhances task decomposition and flexibility, it also amplifies the consequences of misinformation, especially when memory, reasoning, and dialogue co-evolve. Ensuring safety thus requires new protocols for cross-agent trust calibration, message validation, and containment of semantic drift.

\subsection{Cross-Sectional Risk Analysis: Mapping Abilities to Autonomy}

While the previous sections decomposed autonomy-induced risks by agent capabilities, a holistic synthesis is needed to capture how these vulnerabilities interact across autonomy levels. Table~\ref{tab:agent_risk_summary} summarizes the evolution of security risks as agents gain capabilities in perception, cognition, planning, decision-making, and execution from partial (L1) to full autonomy (L5). Each cell encodes a failure mode along with its structural cause ($\leftarrow$) and impact ($\rightarrow$), and is color-coded to reflect severity.

This taxonomy reveals a consistent trend: increasing autonomy does not merely escalate the magnitude of risks—it transforms their structure. For instance, perception errors progress from shallow prompt confusion at L2 to hallucinated internal sensing at L5~\cite{tian2023evil,yuan2019adversarial}. These shifts undermine external observability and traceability, making traditional detection methods ineffective.

In the cognitive dimension, agents at higher levels develop persistent memory and internal reasoning loops. The additions allow for retrospective inference and learning, but also create vulnerabilities such as memory poisoning, biased recall, and value drift~\cite{dong2025practical,Hatalis2024Memory,shinn2023reflexion}. Unlike prompt-level attacks, these threats emerge from within, accumulating through recursive self-reinforcement and violating what we term the \emph{reflective integrity constraint} in CMDP modeling.

Planning capabilities exhibit a similar transition. Early-stage agents misuse fixed scripts, while higher-autonomy agents can generate unsafe plans due to contaminated subgoal chains or self-derived meta-strategies~\cite{Pan2022RewardMisspecification,Farquhar2025MONA,Zhang2024Action}. In CMDP terms, this expands the effective action space and perturbs the transition model $P(s'|s,a)$ with long-horizon dependencies and latent misalignment.

The decision-making dimension underscores the loss of reversibility. Human oversight guarantees safety at L1, but is replaced by heuristic trade-offs at L3, and opaque meta-decisions at L5. These transitions erode accountability, as agent decisions become the product of latent value systems rather than explicit constraints~\cite{Weng2024RewardHacking,Tan2025Reflective}.

Finally, in the execution layer, the risks shift from supervised action to irreversible, autonomous actuation. Tool use, API calls, and system-level actuation introduce high-stakes consequences if left ungoverned~\cite{fu2024imprompt,Yang2024Backdoor}. For agents that continuously act without external halting mechanisms, this creates unstoppable loops—violations of CMDP constraints on bounded action budgets or terminal states.

Taken together, this matrix formalizes the risk landscape of LLM-powered agents and supports a shift from reactive patching toward proactive design. Each risk class corresponds to a type of CMDP constraint violation—be it state consistency, plan safety, policy boundedness, or actuation reversibility. Future agent architectures must encode such constraints as first-class elements, enforced both structurally and temporally, to ensure that as autonomy scales, safety scales with it.

\section{Toward Integrated Safety Architectures in LM-Based Agents}
\label{sec:rhf_architecture}

As large-model agents become increasingly autonomous, ensuring their safety requires more than localized patches or prompt-level filtering. Recent research trends point toward systematically integrated architectures—designs that embed safety considerations across perception, memory, planning, decision-making, and feedback. In this section, we first analyze the technical requirements and architectural shifts driving this trend. We then survey recent defense efforts across functional layers and autonomy levels, highlighting both their contributions and their limitations. Finally, we synthesize these insights into a forward-looking architectural perspective, outlining the principles behind emerging reflective and risk-aware agent designs.

\subsection{Systematizing Risk Mitigation Approaches in LM-Based Agents}

To synthesize the diverse research landscape on the safety of large-model AI agents, we organize prior work around a layered structure that reflects common points of intervention within the agent's decision-making loop. Rather than proposing a novel framework, this view distills how researchers have sought to mitigate failure modes at distinct functional boundaries—ranging from input preprocessing to post-hoc correction. These layers align conceptually with classical theories such as Solomonoff induction~\cite{solomonoff1964formal}, embedded decision theory~\cite{soares2015formal}, and Vingean reflection~\cite{soares2015logical}, while grounding them in the mechanisms specific to foundation model (FM) agents.

At the input boundary, defenses aim to sanitize or reinterpret prompts and sensory signals before they influence internal computations. Recent work on robust multimodal alignment~\cite{lin2023vila}, adversarial training~\cite{chen2024MAT}, and structured rewriting~\cite{chen2024struq} attempts to intercept malicious or ambiguous inputs that may induce cascading failures. These methods form a “semantic firewall” but remain vulnerable to latent threats, such as multi-agent prompt infection~\cite{lee2024prompt} and context drift in long-horizon dialogues~\cite{cobb2022context}.

Within the agent's planning and reasoning phase, risk-aware decision-making introduces formal structure to otherwise opaque inference processes. Constrained MDP-based approaches such as RiskTransformer~\cite{chen2024riskTransformer}, Elastic Decision Transformer~\cite{wu2023edt}, and ARDT~\cite{tang2024ardt} integrate utility optimization with safety constraints. These models simulate alternative trajectories and evaluate their potential for constraint violations, offering partial restoration of classical agent assumptions such as model-based control and goal-directed behavior under uncertainty~\cite{soares2015formal, hubinger2019risks}.

At the execution boundary, proposed actions are filtered through runtime safeguards that verify alignment, detect anomalies, or trigger human oversight. Mechanisms such as process supervision~\cite{luo2024processSupervision}, tool access policies~\cite{qin2024toolllm}, and self-verbalization~\cite{shinn2023reflexion} operate as last-line defenses before the agent interacts with external systems. However, the lack of action explainability and generalized policy transfer limits their scope across complex toolchains.

Finally, learning from feedback represents the most adaptive layer. Approaches such as CALM auditing~\cite{zheng2025calm}, RLAIF~\cite{zhou2023rlaif}, and immune-style learning~\cite{uncknown2024faultHealing} aim to update the agent’s internal objectives and reasoning process over time. Preference-editing systems like PRELUDE~\cite{gao2024prelude} allow users to modify agent behavior via natural language, introducing lightweight post-hoc alignment that avoids full retraining. These reflective mechanisms help agents accumulate a safety-aware prior, improving their ability to anticipate harm, defer uncertain actions, and internalize ethical constraints.

Together, this layered decomposition reveals a structured yet incomplete landscape: while each intervention level addresses specific vulnerabilities, persistent challenges remain in coordination across levels, generalization to novel tasks, and achieving provable guarantees for aligned autonomous behavior.

\subsection{Comprehensive Defense Stack for LM-Based Autonomous Agents}

To ground our discussion of agent safety in concrete operational capabilities, we outline a functional risk progression across five autonomy levels (L1–L5), spanning from partially autonomous tools to fully self-directed agents. Each level reflects a qualitatively distinct degree of agentic independence, as autonomy scales across core functional components—perception, cognition, planning, decision-making, and execution.
While our earlier analysis structured these dimensions in tabular form, the underlying trends can be captured narratively. A key insight emerges: \textbf{as agent autonomy increases, safety requirements shift from procedural safeguards to systemic and architectural guarantees.}

\begin{itemize}
    \item \textbf{L1--L2 (Tool and Consultant Agents):} Risks are largely local and remain addressable via human-in-the-loop oversight, rule-based input filtering, and constrained action spaces. Defense strategies include calibrated sensing, static plan validation, and manual override mechanisms.
    
    \item \textbf{L3 (Collaborator Agents):} Agents possess persistent memory and environment-aware planning. Ensuring safety requires dynamic policy auditing, rollback logic, anomaly detection, and stochastic planning under uncertainty.
    
    \item \textbf{L4--L5 (Quasi-Full and Fully Autonomous Agents):} Operating in open-ended, high-stakes settings, these agents demand end-to-end alignment safeguards. Defense mechanisms include constrained actuation spaces, self-healing perceptual pipelines, policy-grounded ethical reasoning, and formal verification of decision logic.
\end{itemize}

\begin{table*}[htbp]
\centering
\small
\renewcommand{\arraystretch}{1.25}
\setlength{\tabcolsep}{5pt}
\caption{Vertically Integrated Security Stack for Large-Model AI Agents: Confirmed Defenses and Outstanding Gaps }
\vspace{-2ex}
\label{tab:vertical_stack_confirmed_refs}

\begin{tabular}{|>{\centering\arraybackslash}m{3cm}|>{\raggedright\arraybackslash}m{9cm}|>{\raggedright\arraybackslash}m{4.6cm}|}
\hline
\textbf{Layer} & \textbf{Representative Defenses} & \textbf{Outstanding Gaps} \\
\hline

\textbf{L0: Perception \& Input Sanitization} &
Multimodal adversarial training~\cite{chen2024MAT}; robust visual-language alignment (VILA)~\cite{lin2023vila}; structured input sanitization~\cite{zheng2025calm}; unsupervised adversarial fine-tuning (Robust CLIP)~\cite{schlarmann2024robustCLIP}; evaluation of active environment injection~\cite{chen2025evaluating} &
Insufficient defense against prompt-embedded perturbations and real-world sensory spoofing~\cite{jiang2025survey} \\
\hline

\textbf{L1: Representation \& Memory Control} &
Agentic memory design (A-MEM)~\cite{xu2025amem}; memory-efficient structured pruning (Mini-LLM)~\cite{cheng2024mini}; multimodal robustness auditing~\cite{wu2024multimodalRobustness}; immune-style fault healing~\cite{uncknown2024faultHealing}; multiprobe alignment checks (AuditLLM)~\cite{amirizaniani2024auditllm} &
Lack of transparency in memory update policies and unsafe state persistence across sessions~\cite{jiang2025survey} \\
\hline

\textbf{L2: Risk-Aware Planning (CMDP)} &
RiskTransformer~\cite{chen2024riskTransformer}; Adversarially Robust Decision Transformer~\cite{tang2024ardt}; Elastic Decision Transformer~\cite{wu2023edt}; RiskQ for multi-agent RL~\cite{sun2023riskq}; self-challenging agents~\cite{zhou2025self} &
Stable planning under non-stationary dynamics and noisy reward signals remains under-addressed \\
\hline

\textbf{L3: Execution Guarding} &
ToolLLM for API grounding~\cite{qin2023toolllm}; process supervision to enforce intermediate alignment~\cite{luo2024processSupervision}; autoregressive actuation with Reflexion~\cite{shinn2023reflexion} &
Runtime enforcement of execution constraints across toolchains remains fragile \\
\hline

\textbf{L4: Feedback \& Self-Governance} &
CALM auditing for curiosity-based feedback~\cite{zheng2025calm}; RLAIF for reward hacking mitigation~\cite{zhou2023rlaif}; immune learning dynamics~\cite{uncknown2024faultHealing}; self-challenging feedback loops~\cite{zhou2025self} &
Handling adversarial and conflicting feedback in open-ended learning loops~\cite{jiang2025survey} \\
\hline

\textbf{Cross-Layer Integration} &
Self-healing agents~\cite{uncknown2024faultHealing}; multimodal memory auditing~\cite{wu2024multimodalRobustness}; compositional risk modeling with RiskTransformer~\cite{chen2024riskTransformer} &
Missing standardized protocols for secure semantic propagation across modules \\
\hline

\end{tabular}
  \vspace{-3ex}
\end{table*}

\subsubsection{\textbf{L0: Perception and Input Sanitization}}

The perception layer acts as the first line of defense in large model-based autonomous agents, converting raw sensory inputs into structured internal representations. It encompasses low-level signal interpretation (e.g., image, audio, text) and early multimodal fusion. Errors at this stage—such as misclassified visuals or ambiguous instructions—can silently propagate through the pipeline, causing semantic drift, reasoning failures, or unsafe actuation.
To mitigate such risks, recent techniques include \emph{multimodal adversarial training}~\cite{chen2024MAT}, \emph{visual-language alignment} (VILA)~\cite{lin2023vila}, and \emph{structured input sanitization}~\cite{zheng2025calm}. Robustness is further improved via \emph{unsupervised adversarial fine-tuning} (e.g., Robust CLIP~\cite{schlarmann2024robustCLIP}) and \emph{environment injection testing}~\cite{chen2025evaluating}.

Yet persistent gaps remain. \textbf{Prompt-embedded attacks}—where malicious payloads are subtly encoded in multimodal inputs—can bypass surface filters and trigger latent misbehavior. \textbf{Real-world spoofing}, such as adversarial stickers or inaudible voice commands, exploits weak invariance in sensory processing~\cite{jiang2025survey}. These vulnerabilities compromise safety-critical constraints: a misperceived pedestrian may yield an otherwise disallowed action, even under a well-aligned policy.
Thus, perception must be not only robust but also \emph{constraint-preserving}, ensuring that derived representations remain safe to act upon throughout the agent’s decision loop.

\subsubsection{\textbf{L1: Representation \& Memory Control}}

This layer manages how agents encode, store, and retrieve knowledge across time. Unlike stateless models, autonomous agents rely on persistent memory to support long-horizon planning, context tracking, and adaptive behavior. However, latent memory artifacts—once corrupted—can reinforce hallucinations, entrench outdated beliefs, or propagate adversarially injected content across episodes.
Recent advances include \emph{schema-sensitive memory} (e.g., A-MEM~\cite{xu2025amem}), \emph{pruned LLMs} for representation control~\cite{cheng2024mini}, \emph{multimodal consistency auditing}~\cite{wu2024multimodalRobustness}, and \emph{immune-style fault healing}~\cite{uncknown2024faultHealing}. Tools like \emph{AuditLLM}~\cite{amirizaniani2024auditllm} probe memory for latent misalignment or unsafe goal traces.

Nonetheless, key concerns persist. Most systems lack visibility into memory updates across session boundaries, with limited guarantees on the deprecation of unsafe or stale information~\cite{jiang2025survey}. Memory corruption is often silent yet persistent, shaping future decisions in harmful ways. For example, an agent that misremembers a prior success with a risky tool may re-execute it in an unsuitable context.
Effective memory control must therefore combine representational robustness with \emph{constraint-consistent persistence}, ensuring that stored knowledge does not silently drift from aligned behavior over time.

\subsubsection{\textbf{L2: Risk-Aware Planning and Decision Logic}}

The planning layer governs how agents select actions over extended time horizons, often under uncertainty and dynamic goals. In large model-based agents, this involves transforming perceptions and internal memory into coherent action sequences using learned or inferred value structures. The introduction of Constrained Markov Decision Process (CMDP) frameworks~\cite{altman1999constrained, chen2024riskTransformer} reflects an important step toward embedding formal safety constraints into the planning core.
Recent works such as \emph{RiskTransformer}~\cite{chen2024riskTransformer} and \emph{Adversarially Robust Decision Transformer (ARDT)}~\cite{tang2024ardt} integrate risk-awareness directly into the policy learning process, allowing agents to weigh expected reward against constraint violation probabilities. \emph{Elastic Decision Transformer}~\cite{wu2023edt} introduces flexible reward modeling to adapt to distribution shifts, while \emph{RiskQ}~\cite{sun2023riskq} extends these ideas to multi-agent reinforcement learning with competitive or cooperative dynamics. Self-corrective agents~\cite{zhou2025self} further embed meta-level checks that monitor plan coherence and constraint adherence over time.

However, core vulnerabilities persist. \textbf{Stability under non-stationary dynamics} remains a major challenge: agents trained in fixed environments often generalize poorly when environmental or goal distributions change. Moreover, reward signals—either learned or externally defined—may be noisy, ambiguous, or misaligned with stakeholder expectations, leading to goal mis-specification or reward hacking. Even in CMDP-based models, safety constraints are often approximated rather than guaranteed, especially under high-dimensional input and stochastic transitions.
To preserve safety, planning must be not only optimal but also \emph{constraint-compliant and risk-calibrated}. Agents should favor strategies that defer unsafe commitments, include rollback mechanisms, and allow for online anomaly detection. Yet, current models largely lack principled tools to quantify or communicate such risk margins during inference.

\subsubsection{\textbf{L3: Execution Guarding and Tool Actuation}}

The execution layer translates high-level plans into concrete actions via external tools, APIs, or physical actuators. For large-model agents interfacing with the real world or digital infrastructure, this marks the point where symbolic errors can result in irreversible consequences.
Key defense efforts focus on maintaining alignment between the model’s internal intent and the external effects of its actions. \emph{ToolLLM}~\cite{qin2023toolllm} grounds language-model predictions in structured tool APIs, reducing ambiguity in execution targets. \emph{Process supervision frameworks}~\cite{luo2024processSupervision} monitor intermediate execution steps to detect divergence from expected plans. \emph{Autoregressive actuation strategies}, such as Reflexion~\cite{shinn2023reflexion}, allow agents to reevaluate and adapt their actions based on ongoing outcomes and self-feedback.

Yet, execution remains one of the most fragile stages in the agent stack. \textbf{Runtime enforcement of constraints across toolchains} is still underdeveloped. Many agents lack transactional safeguards, sandboxed execution environments, or robust rollbacks. This creates a risk surface where single-step hallucinations yield system-level failures—especially in high-privilege or high-impact contexts such as financial APIs, robotics control, or medical decision systems.
To improve resilience, execution must be both \emph{transparent} and \emph{recoverable}. Action traces should be auditable post-hoc, and failure cascades should be interrupted by safety interlocks. However, these safeguards are rarely implemented or formally verified in current agent architectures.

\subsubsection{\textbf{L4: Feedback and Self-Governance}}

Feedback mechanisms are essential for long-term adaptation and value alignment in autonomous agents. At this layer, the agent learns from its environment, users, or internal metrics—modifying its goals, beliefs, or behaviors accordingly. This feedback can be implicit (e.g., environment reward), explicit (e.g., user correction), or introspective (e.g., policy reflection). \emph{Reward Learning from AI Feedback (RLAIF)}~\cite{zhou2023rlaif} and \emph{CALM auditing}~\cite{zheng2025calm} introduce structured ways to evaluate whether agent behavior aligns with intended ethical or task objectives. \emph{Self-challenging loops}~\cite{zhou2025self} allow agents to proactively generate counterexamples to their own strategies, enhancing robustness to value drift. Immune-style learning~\cite{uncknown2024faultHealing} further proposes anomaly-driven feedback mechanisms inspired by biological systems.

Nonetheless, this layer faces deep challenges. Feedback channels are vulnerable to \textbf{manipulation, ambiguity, or misinterpretation}. In open-ended learning, agents may reinforce unsafe behaviors if the feedback signal is biased or adversarial~\cite{jiang2025survey}. Furthermore, high-autonomy agents may generate their own learning signals, creating the risk of self-confirming value loops or deceptive alignment, where apparent compliance masks divergent objectives. Effective feedback processing must be \emph{multi-perspective and self-critical}, not merely reward-maximizing. Agents should be able to reason over conflicting value inputs, differentiate short-term performance from long-term safety, and incorporate negative feedback without destabilizing their internal state. Current systems are far from achieving such reflexive governance in practice.

\subsubsection{Limitations and Discussion}

The recent rise of large-model-based autonomous agents reveals a core tension among three foundational objectives: \textit{autonomy}, \textit{controllability}, and \textit{robustness}. As agents acquire increasingly general capabilities—long-horizon planning, dynamic tool use, and self-reflection—external oversight becomes weaker, and behavioral auditing more complex. Simultaneously, improvements in robustness often introduce architectural opacity, diminishing transparency and complicating alignment with human intent.
This autonomy–controllability–robustness triad underlies many emergent failure modes: brittle generalization, goal misalignment, unsafe reward optimization, and unbounded tool invocation. As agents cross the L2 autonomy threshold, static constraints and manual intervention become unsustainable for ensuring safety.

However, most existing LLM-agent frameworks fail to meet these demands. They often rely on reactive, prompt-driven reasoning without an explicit forward model, making it difficult for agents to anticipate the consequences of their actions. Reward signals are typically embedded in task heuristics or implicit success markers, which blurs the distinction between utility modeling and behavioral constraint. Moreover, risk is rarely treated as a first-class modeling objective—agents lack internal mechanisms for forecasting potential hazards or enforcing safety budgets. Although some frameworks incorporate forms of self-reflection, they are often confined to linguistic output and disconnected from actual policy correction.

\subsection{Reflective Risk-Aware Agent Architecture}

\begin{figure}[!t]
    \centering
    \includegraphics[width=1\linewidth]{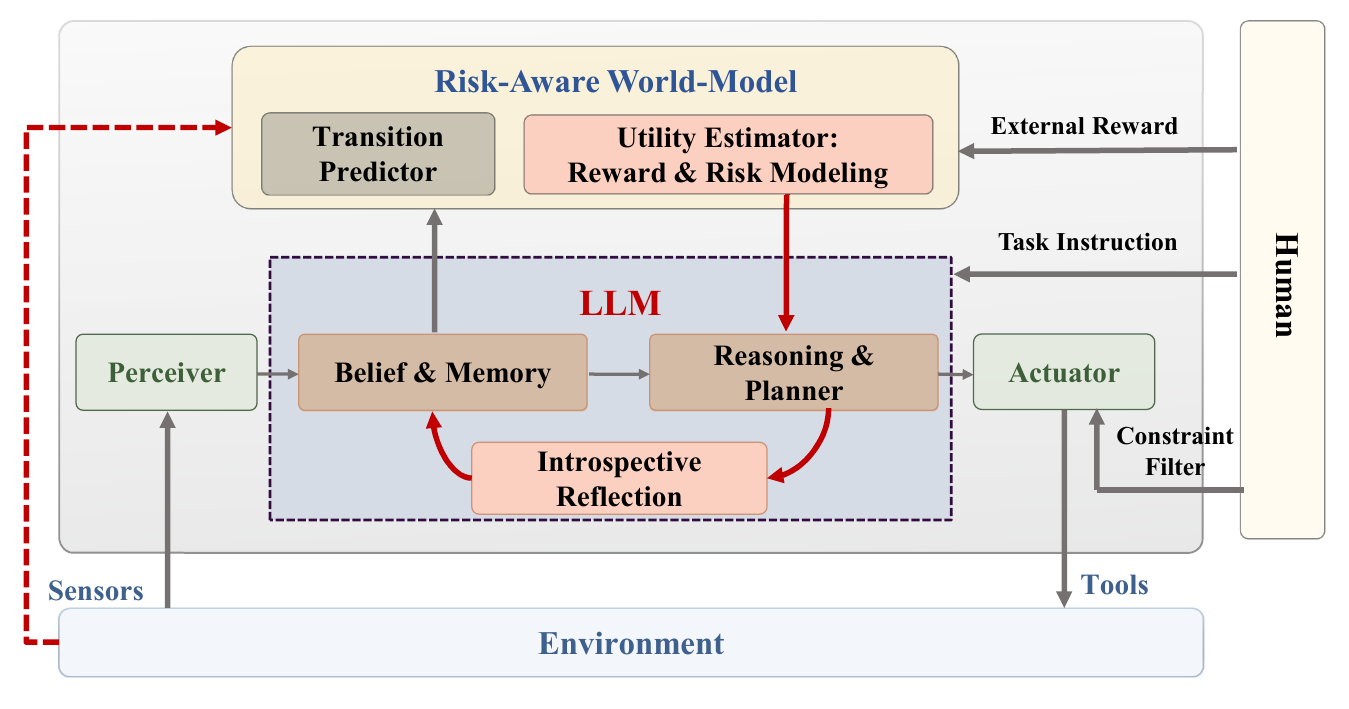}
      \vspace{-5ex}
    \caption{ The \textbf{Reflective Risk-Aware Agent Architecture (R2A2)}. The agent integrates a large language model (LLM)-driven reasoning core with a risk-aware world model for long-horizon planning under constraints. Perceptual inputs are processed into latent state representations by the \texttt{Perceiver} and maintained in \texttt{Belief \& Memory}. The LLM generates actions via the \texttt{Reasoning \& Planner} module, which are evaluated by the \texttt{Utility Estimator} for reward and risk. Unsafe actions are blocked by a \texttt{Constraint Filter}, while policy refinement occurs via \texttt{Introspective Reflection}. Human instructions and rewards modulate utility estimation and planning.}
      \vspace{-2ex}
    \label{fig:R2A2_Arch}
\end{figure}

To address these structural limitations, we propose the \textit{Reflective Risk-Aware Agent Architecture (R2A2)}—a modular framework that integrates predictive simulation, utility estimation, constraint filtering, and introspective reflection. Each component is grounded in the semantics of constrained sequential decision-making.
Fig.~\ref{fig:R2A2_Arch} illustrates the architecture of R2A2, which supports reflective planning, risk-aware reasoning, and constraint-sensitive execution for LLM-based agents.

Perceptual signals from the environment are processed by a \texttt{Perceiver} and encoded into latent state via the \texttt{Belief \& Memory} module. These internal representations are passed to the LLM, which drives both the \texttt{Reasoning \& Planner} and \texttt{Introspective Reflection} modules. While the planner generates action proposals aligned with the task, the reflection module retrospectively analyzes behavior and contradictions to adapt policy trajectories over time.
R2A2 introduces a dedicated \texttt{Risk-Aware World Model}, composed of a \texttt{Transition Predictor} and a joint \texttt{Utility Estimator}. These simulate future states and evaluate both rewards and risks associated with candidate actions, enabling risk-sensitive planning. Proposed actions are screened through a \texttt{Constraint Filter}, which blocks unsafe plans before execution by the \texttt{Actuator}. Human supervision is incorporated through task instructions and reward feedback, which influence both belief updating and utility modeling.

As large-model agents gain autonomy—capable of long-horizon planning, dynamic tool use, and adaptive behavior—they must balance performance with safety across multiple dimensions. Traditional Markov Decision Processes (MDPs) offer no explicit mechanism to encode operational risks, such as privacy leakage, irreversible actions, or resource overuse. Conversely, heuristic rule filters or ad hoc safety checks provide limited generalization and no guarantees.
The \textit{Constrained Markov Decision Process} (CMDP) framework~\cite{altman1999constrained} offers a principled foundation for reasoning under both reward and constraint. By augmenting the standard MDP with explicit cost functions and budget thresholds, CMDPs enable agents to optimize long-term utility while respecting multi-dimensional safety bounds. The agent seeks a policy $\pi^*$ that maximizes expected cumulative reward while satisfying multiple risk constraints:
\begin{align}
\pi^{*} \;=\; 
\arg\max_{\pi}\;
        \mathbb{E}_{\pi}\!\Big[\!\sum_{t=0}^{\infty}\!\gamma^{t} R(s_t,a_t)\Big] \nonumber \\
\quad\text{s.t.}\quad 
        \mathbb{E}_{\pi}\!\Big[\!\sum_{t=0}^{\infty}\!\gamma^{t} C_k(s_t,a_t)\Big]
        \le d_k,\;\forall k. \label{eq:cmdp}
\end{align}

To operationalize this objective, R2A2 employs a primal–dual control mechanism that selects actions by balancing reward and constraint satisfaction via Lagrangian relaxation. At each step, the agent chooses the optimal action $a_t^*$ and updates its dual variables $\lambda_t$ as follows:
\begin{align}
a_t^{*} 
&= \arg\max_{a}\;
        Q_R(s_t,a)\;+\;\langle \lambda_t,\; d-Q_C(s_t,a)\rangle ,\nonumber\\[2pt]
\lambda_{t+1} 
&= \bigl[\lambda_t\;+\;\eta\bigl(Q_C(s_t,a_t^{*})-d\bigr)\bigr]_{+}, \label{eq:primal_dual}
\end{align}
where $Q_R$ and $Q_C$ denote the estimated reward and cumulative cost, $\eta$ is the dual learning rate, and $[\cdot]_+$ projects onto the non-negative orthant. This mechanism enables adaptive risk-sensitive planning, and under standard conditions (e.g., bounded cost, Lipschitz continuity), it converges to an $\varepsilon$-feasible policy~\cite{paternain2019constrained}, providing quantitative safety guarantees for real-world deployment.

\textbf{Safety-Driven Design Discussion.}
The Reflective Risk-Aware Agent Architecture (R2A2) addresses the increasingly critical safety challenges posed by highly autonomous large-model agents, including interactional autonomy, continuous learning drift, emergent goal generation, irreversible external impacts, unbounded resource access, and unpredictable alignment behavior. These risks, largely entangled and implicit in prior frameworks such as Weng’s LLM agent stack~\cite{weng2023safe}, are explicitly modeled and mitigated in R2A2 through a tightly integrated architectural design grounded in Constrained Markov Decision Processes (CMDPs).

Specifically, R2A2 introduces a cohesive safety stack that integrates three core capabilities. First, a \textbf{structured reasoning–reflection loop} allows the agent to generate plans via goal-conditioned chain-of-thought reasoning, and to retrospectively analyze its own decisions for logical inconsistency, constraint violations, or value misalignment. This mechanism acts as an internal audit process, offering adaptive correction beyond static rules or prompt filtering. Second, a \textbf{unified utility estimator} jointly models reward and multi-dimensional risks within a shared optimization space, enabling trade-off-aware planning under strict safety budgets. Unlike heuristic or entangled scoring in prior systems, this module enforces explicit bounds on behavior through constraint filtering. Third, a \textbf{forward-looking simulation path}, powered by the \texttt{Transition Predictor}, allows the agent to forecast the effects of its actions before execution—effectively closing the loop between planning and consequence evaluation. By modeling all planning and action selection within the CMDP framework, R2A2 not only ensures reward-optimality under cost constraints but also provides a formal scaffold for introspection, adaptation, and safety auditing.

\section{Conclusion and Future Work}
\label{Conclusion}

The emergence of large-model autonomous agents marks a fundamental shift in AI system design—from static predictors to interactive, tool-using, and self-reflective decision-makers. While these agents offer unprecedented flexibility across domains, they also introduce qualitatively new vulnerabilities that span memory, planning, tool use, and emergent behavior. As we have shown, the autonomy of such agents expands their capability frontier, but simultaneously reshapes the threat landscape, making traditional safety measures insufficient.

Underlying many of these vulnerabilities is a core design tension among three foundational dimensions: \textbf{autonomy}, \textbf{controllability}, and \textbf{robustness}. This trade-off triangle defines the boundaries of safe agent behavior. As agents become more autonomous—planning over long horizons and acting without human intervention—their behavior becomes harder to interpret, verify, or override, thus diminishing controllability. Conversely, enhancing robustness to adversarial perturbations or distribution shifts often entails architectural complexity and reduced transparency, which can further impede oversight. Failure modes such as delegation breakdowns, deceptive reflection, or brittle generalization often emerge at the edges of this triangle, where one capability is optimized at the cost of the others.

To address this, we proposed the \textbf{Reflective Risk-Aware Agent Architecture (R2A2)}, a unified design framework that embeds safety, alignment, and risk-awareness into the agent's cognitive loop. By modeling the agent’s decision process through a CMDP formulation and implementing layered safeguards—including threat-conditioned planning, modular policy enforcement, and dual-reward arbitration—R2A2 promotes endogenous security: agents that reason about risks, anticipate failures, and adapt behavior in alignment with human values.
As foundation model agents continue to evolve, resolving the autonomy–controllability–robustness trade-off will be central to building AI systems we can trust.

\subsection{Future Work}

Our paper highlights that the increasing autonomy and complexity of LLM-based agents introduces both significant opportunities and unresolved safety challenges. Based on the current research landscape, we outline several promising directions for future investigation.

First, a central open problem is the development of \textbf{verifiable safety guarantees} for high-capacity agents. While constrained decision-making frameworks such as CMDPs offer formal abstractions, scalable and expressive verification methods remain limited—especially for policies conditioned on high-dimensional, symbolic language inputs. Recent efforts such as ShieldAgent~\cite{shieldagent2025} and formal-method-integrated frameworks~\cite{zhang2024fusion} aim to bridge this gap by combining logic-based safety policies with verifiable execution traces, but scalable integration with learning-based agents remains a challenge.

Second, there is growing consensus on the need for \textbf{full-stack evaluation environments} beyond prompt-level perturbations. ShieldAgent-Bench~\cite{shieldagent2025} and MLA‑Trust~\cite{yang2025mla} exemplify this shift by offering risk-labeled agent trajectories and multimodal, tool-interacting benchmarks, respectively. These richer datasets allow for assessment of safety over persistent memory, API usage, and long-horizon planning—uncovering delayed or compounding failure modes difficult to capture via static input-output testing.

Third, advances in \textbf{reflective self-monitoring and feedback integration} are likely to define the next frontier of agent safety. Works such as STAIR~\cite{zhang2025stair}, RealSafe-R1~\cite{zhang2025realsafe}, Re-ReST~\cite{rerefst2024} and AegisLLM~\cite{aegisllm2025} demonstrate the potential of introspection-enhanced learning and runtime guard agents that adaptively respond to emergent threats. Future agents may benefit from combining self-reflective reasoning with user clarification queries or sentinel feedback from the environment to proactively detect and mitigate risks during deployment.

Finally, as multi-agent systems proliferate, understanding \textbf{emergent risks in collective agent behavior} is increasingly critical. Works like Reflective Multi-Agent Collaboration (COPPER)~\cite{copper2024} and AegisLLM~\cite{aegisllm2025} explore intra-agent introspection and coordination guardrails. Additionally, the newly proposed MAEBE framework~\cite{maebe2025} offers a benchmarked evaluation of emergent behavior risks—such as peer pressure and alignment divergence—in agent ensembles. Future safety research must develop game-theoretic safeguards, decentralized trust models, and population-level formal analyses to ensure robust coordination and safety in open agent ecologies.


%\section*{References}
{
\bibliographystyle{IEEEtran}
\bibliography{references}

\begin{thebibliography}{100}
\providecommand{\url}[1]{#1}
\csname url@samestyle\endcsname
\providecommand{\newblock}{\relax}
\providecommand{\bibinfo}[2]{#2}
\providecommand{\BIBentrySTDinterwordspacing}{\spaceskip=0pt\relax}
\providecommand{\BIBentryALTinterwordstretchfactor}{4}
\providecommand{\BIBentryALTinterwordspacing}{\spaceskip=\fontdimen2\font plus
\BIBentryALTinterwordstretchfactor\fontdimen3\font minus \fontdimen4\font\relax}
\providecommand{\BIBforeignlanguage}[2]{{%
\expandafter\ifx\csname l@#1\endcsname\relax
\typeout{** WARNING: IEEEtran.bst: No hyphenation pattern has been}%
\typeout{** loaded for the language `#1'. Using the pattern for}%
\typeout{** the default language instead.}%
\else
\language=\csname l@#1\endcsname
\fi
#2}}
\providecommand{\BIBdecl}{\relax}
\BIBdecl

\bibitem{sapkota2025ai}
R.~Sapkota, K.~I. Roumeliotis, and M.~Karkee, ``{{AI}} agents vs. agentic {{AI}}: A conceptual taxonomy, applications and challenge,'' \emph{arXiv preprint arXiv:2505.10468}, 2025.

\bibitem{Xi2023Rise}
Z.~Xi, W.~Chen, X.~Guo, W.~He, Y.~Ding, B.~Hong, M.~Zhang, J.~Wang, S.~Jin, E.~Zhou, R.~Zheng, X.~Fan, X.~Wang, L.~Xiong, Y.~Zhou, W.~Wang, C.~Jiang, Y.~Zou, X.~Liu, Z.~Yin, S.~Dou, R.~Weng, W.~Cheng, Q.~Zhang, W.~Qin, Y.~Zheng, X.~Qiu, X.~Huang, and T.~Gui, ``The rise and potential of large language model based agents: A survey,'' \emph{arXiv preprint arXiv:2309.07864}, 2023.

\bibitem{bommasani2021opportunities}
R.~Bommasani, D.~A. Hudson, E.~Adeli, R.~Altman, S.~Arora, S.~von Arx, M.~S. Bernstein, J.~Bohg, A.~Bosselut, E.~Brunskill \emph{et~al.}, ``On the opportunities and risks of foundation models,'' \emph{arXiv preprint arXiv:2108.07258}, 2021.

\bibitem{openai2023gpt}
OpenAI, ``{{GPT}}-4 technical report,'' \emph{arXiv preprint arXiv:2303.08774}, 2023.

\bibitem{grattafiori2024llama}
A.~Grattafiori, A.~Dubey, A.~Jauhri, A.~Pandey, A.~Kadian, A.~Al-Dahle, A.~Letman, A.~Mathur, A.~Schelten, A.~Vaughan \emph{et~al.}, ``The llama 3 herd of models,'' \emph{arXiv preprint arXiv:2407.21783}, 2024.

\bibitem{yao2023react}
S.~Yao, J.~Zhao, D.~Yu, N.~Du, I.~Shafran, K.~Narasimhan, and Y.~Cao, ``React: Synergizing reasoning and acting in language models,'' in \emph{International Conference on Learning Representations}, 2023.

\bibitem{schick2023toolformer}
T.~Schick, J.~Dwivedi-Yu, R.~Dess{\`\i}, R.~Raileanu, M.~Lomeli, E.~Hambro, L.~Zettlemoyer, N.~Cancedda, and T.~Scialom, ``Toolformer: Language models can teach themselves to use tools,'' \emph{Advances in Neural Information Processing Systems}, vol.~36, pp. 68\,539--68\,551, 2023.

\bibitem{Nakano2021WebGPT}
R.~Nakano, J.~Hilton, S.~Balaji, L.~Ouyang, J.~Wu, C.~Kim, C.~Hesse, S.~Jain, V.~Kosaraju, W.~Saunders, X.~Jiang, K.~Cobbe, T.~Eloundou, G.~Krueger, K.~Button, M.~Knight, B.~Chess, and J.~Schulman, ``Webgpt: Browser-assisted question-answering with human feedback,'' \emph{arXiv preprint arXiv:2112.09332}, 2021.

\bibitem{wang2023voyager}
G.~Wang, Y.~Xie, Y.~Jiang, A.~Mandlekar, C.~Xiao, Y.~Zhu, L.~Fan, and A.~Anandkumar, ``Voyager: An open-ended embodied agent with large language models,'' \emph{arXiv preprint arXiv:2305.16291}, 2023.

\bibitem{yang2024AutoGPT}
H.~Yang, Y.~Sifu, and Y.~He, ``Auto-{{GPT}} for online decision making: Benchmarks and additional opinions,'' \emph{arXiv preprint arXiv:2306.02224}, 2023.

\bibitem{he2024security}
Y.~He, E.~Wang, Y.~Rong, Z.~Cheng, and H.~Chen, ``Security of {{AI}} agents,'' \emph{arXiv preprint arXiv:2406.08689}, 2024.

\bibitem{wu2023autogen}
Q.~Wu, G.~Bansal, J.~Zhang, Y.~Wu, B.~Li, E.~Zhu, L.~Jiang, X.~Zhang, S.~Zhang, J.~Liu \emph{et~al.}, ``Autogen: Enabling next-gen llm applications via multi-agent conversation,'' \emph{arXiv preprint arXiv:2308.08155}, 2023.

\bibitem{russell2021ai}
S.~Russell and P.~Norvig, \emph{Artificial Intelligence: A Modern Approach}, 4th~ed.\hskip 1em plus 0.5em minus 0.4em\relax Pearson, 2021.

\bibitem{dunzelman2023comprehensive}
N.~Dunzelman, ``A comprehensive review of autonomous agents powered by large language models,'' \emph{Medium}, 2023.

\bibitem{wang2023survey}
L.~Wang, C.~Ma, X.~Feng, Z.~Zhang, H.~Yang, J.~Zhang, Z.~Chen, J.~Tang, X.~Chen, Y.~Lin, W.~X. Zhao, Z.~Wei, and J.-R. Wen, ``A survey on large language model based autonomous agents,'' \emph{arXiv preprint arXiv:2308.11432}, 2023.

\bibitem{luo2025survey}
J.~Luo, W.~Zhang, and Y.~e.~a. Yuan, ``Large language model agent: A survey on methodology, applications and challenges,'' \emph{Nature Machine Intelligence}, 2025.

\bibitem{bostrom2014superintelligence}
N.~Bostrom, \emph{Superintelligence: Paths, Dangers, Strategies}.\hskip 1em plus 0.5em minus 0.4em\relax Oxford University Press, 2014.

\bibitem{bengio2024extremeAIRisks}
Y.~Bengio, G.~Hinton, A.~Yao, D.~Song, P.~Abbeel, T.~Darrell, Y.~N. Harari, Y.-Q. Zhang, L.~Xue, S.~Shalev-Shwartz, G.~Hadfield, J.~Clune, T.~Maharaj, F.~Hutter, A.~G. Baydin, S.~McIlraith, Q.~Gao, A.~Acharya, D.~Krueger, A.~Dragan, P.~Torr, S.~Russell, D.~Kahneman, J.~Brauner, and S.~Mindermann, ``{Managing Extreme AI Risks Amid Rapid Progress},'' \emph{Science}, vol. 384, no. 6698, pp. 842--845, 2024.

\bibitem{muehlhauser2012intelligence}
L.~Muehlhauser and A.~Salamon, ``Intelligence explosion: Evidence and import,'' \emph{In Singularity Hypotheses}, pp. 15--42, 2012.

\bibitem{omohundro2018basic}
S.~M. Omohundro, ``The basic ai drives,'' in \emph{Artificial intelligence safety and security}, 2018, pp. 47--55.

\bibitem{bengio2025scientistai}
Y.~Bengio, M.~Cohen, D.~Fornasiere, J.~Ghosn, P.~Greiner, M.~MacDermott, S.~Mindermann, A.~Oberman, J.~Richardson, O.~Richardson, M.-A. Rondeau, P.-L. St-Charles, and D.~Williams-King, ``{Superintelligent Agents Pose Catastrophic Risks: Can Scientist AI Offer a Safer Path?}'' \emph{arXiv preprint arXiv:2502.15657}, 2025.

\bibitem{bengio2024aisafetyreport}
Y.~e.~a. Bengio, ``{International Scientific Report on the Safety of Advanced AI: Interim Report},'' \emph{arXiv preprint arXiv:2412.05282}, 2024.

\bibitem{Borghoff2025Human}
U.~M. Borghoff, P.~Bottoni, and R.~Pareschi, ``Human-artificial interaction in the age of agentic ai: a system-theoretical approach,'' \emph{Frontiers in Human Dynamics}, vol.~7, p. 1579166, 2025.

\bibitem{soares2017agent}
N.~Soares and B.~Fallenstein, ``Agent foundations for aligning machine intelligence with human interests: a technical research agenda,'' \emph{The technological singularity: Managing the journey}, pp. 103--125, 2017.

\bibitem{wang2023describe}
Z.~Wang, S.~Cai, G.~Chen, A.~Liu, X.~Ma, and Y.~Liang, ``Describe, explain, plan and select: Interactive planning with large language models enables open-world multi-task agents,'' \emph{arXiv preprint arXiv:2302.01560}, 2023.

\bibitem{deng2024aiagent}
Z.~Deng, Y.~Guo, C.~Han, W.~Ma, J.~Xiong, S.~Wen, and Y.~Xiang, ``{{AI}} agent security: Emerging threats and defense paradigms,'' \emph{arXiv preprint arXiv:2408.29007}, 2024.

\bibitem{zhang2025advlm}
T.~Zhang, L.~Wang, X.~Zhang, Y.~Zhang, B.~Jia, S.~Liang, S.~Hu, Q.~Fu, A.~Liu, and X.~Liu, ``{{AdvLM}}: Visual adversarial attacks on autonomous driving vision-language models,'' in \emph{Proceedings of the IEEE/CVF Conference on Computer Vision and Pattern Recognition}, 2025, pp. 11\,289--11\,300.

\bibitem{zhang2024agent}
H.~Zhang, J.~Huang, K.~Mei, Y.~Yao, Z.~Wang, C.~Zhan, H.~Wang, and Y.~Zhang, ``{{Agent Security Bench (ASB)}}: Formalizing and benchmarking attacks and defenses in {{LLM}}-based agents,'' \emph{arXiv preprint arXiv:2410.02644}, 2024.

\bibitem{li2025asb}
C.~Li, H.~Wang, W.~Zhang, and M.~Liu, ``{{ASB}}: A security benchmark for {{LLM}}-based autonomous agents,'' \emph{IEEE Transactions on Dependable and Secure Computing}, vol.~22, no.~3, pp. 1451--1468, 2025.

\bibitem{zhang2024breaking}
B.~Zhang, Y.~Tan, Y.~Shen, A.~Salem, M.~Backes, S.~Zannettou, and Y.~Zhang, ``Breaking agents: Compromising autonomous {{LLM}} agents through malfunction amplification,'' \emph{arXiv preprint arXiv:2407.20859}, 2024.

\bibitem{Hua2024TrustAgent}
W.~Hua, X.~Yang, M.~Jin, Z.~Li, W.~Cheng, R.~Tang, and Y.~Zhang, ``{{TrustAgent}}: Towards safe and trustworthy {{LLM}}-based agents,'' \emph{Findings of the Association for Computational Linguistics: EMNLP}, 2024.

\bibitem{yu2025survey}
M.~Yu, F.~Meng, X.~Zhou, S.~Wang, J.~Mao, L.~Pang, T.~Chen, K.~Wang, X.~Li, Y.~Zhang \emph{et~al.}, ``A survey on trustworthy llm agents: Threats and countermeasures,'' \emph{arXiv preprint arXiv:2503.09648}, 2025.

\bibitem{li2025commercial}
A.~Li, Y.~Zhou, V.~C. Raghuram, T.~Goldstein, and M.~Goldblum, ``Commercial llm agents are already vulnerable to simple yet dangerous attacks,'' \emph{arXiv preprint arXiv:2502.08586}, 2025.

\bibitem{chen2025obvious}
C.~Chen, Z.~Zhang, B.~Guo, S.~Ma, I.~Khalilov, S.~A. Gebreegziabher, Y.~Ye, Z.~Xiao, Y.~Yao, T.~Li \emph{et~al.}, ``The obvious invisible threat: Llm-powered gui agents' vulnerability to fine-print injections,'' \emph{arXiv preprint arXiv:2504.11281}, 2025.

\bibitem{narajala2025securing}
V.~S. Narajala and O.~Narayan, ``Securing agentic ai: A comprehensive threat model and mitigation framework for generative ai agents,'' \emph{arXiv preprint arXiv:2504.19956}, 2025.

\bibitem{cheng2024exploring}
Y.~Cheng, C.~Zhang, Z.~Zhang, X.~Meng, S.~Hong, W.~Li, Z.~Wang, Z.~Wang, F.~Yin, J.~Zhao \emph{et~al.}, ``Exploring large language model based intelligent agents: Definitions, methods, and prospects,'' \emph{arXiv preprint arXiv:2401.03428}, 2024.

\bibitem{yang2025mla}
X.~Yang, J.~Chen, J.~Luo, Z.~Fang, Y.~Dong, H.~Su, and J.~Zhu, ``Mla‑trust: Benchmarking trustworthiness of multimodal llm agents in gui environments,'' \emph{arXiv preprint arXiv:2506.01616}, 2025.

\bibitem{openai2024alignment}
O.~S. Team, ``Alignment challenges in autonomous language agents,'' \emph{arXiv preprint arXiv:2403.17815}, 2024.

\bibitem{ma2024caution}
X.~Ma, Y.~Wang, Y.~Yao, T.~Yuan, A.~Zhang, Z.~Zhang, and H.~Zhao, ``Caution for the environment: Multimodal agents are susceptible to environmental distractions,'' \emph{arXiv preprint arXiv:2408.02544}, 2024.

\bibitem{Debenedetti2024AgentDojo}
E.~Debenedetti \emph{et~al.}, ``{{AgentDojo}}: A dynamic environment to evaluate prompt injection attacks and defenses for {{LLM}} agents,'' \emph{arXiv preprint arXiv:2406.13352}, 2024.

\bibitem{Zhang2024Action}
Y.~Zhang, K.~Chen, X.~Jiang, Y.~Sun, R.~Wang, and L.~Wang, ``Towards action hijacking of large language model-based agent,'' \emph{arXiv preprint arXiv:2412.10807}, 2024.

\bibitem{chen2024agentpoison}
Z.~Chen, Z.~Xiang, C.~Xiao, D.~Song, and B.~Li, ``{{AgentPoison}}: Red-teaming {{LLM}} agents via poisoning memory or knowledge bases,'' \emph{Advances in Neural Information Processing Systems}, vol.~37, pp. 130\,185--130\,213, 2024.

\bibitem{Tan2025Reflective}
Z.~Tan, J.~Yan, I.-H. Hsu, R.~Han, Z.~Wang, L.~T. Le, Y.~Song, Y.~Chen, H.~Palangi, G.~Lee, A.~Iyer, T.~Chen, H.~Liu, C.-Y. Lee, and T.~Pfister, ``In prospect and retrospect: Reflective memory management for long-term personalized dialogue agents,'' \emph{arXiv preprint arXiv:2503.08026}, 2025.

\bibitem{qu2024exploration}
C.~Qu, S.~Dai, X.~Wei, H.~Cai, and S.~Wu, ``From exploration to mastery: Enabling {{LLMs}} to master tools via self-driven interactions,'' \emph{arXiv preprint arXiv:2410.08197}, 2024.

\bibitem{Podar2025}
H.~Podar and A.~Colijn, ``Technical risks of (lethal) autonomous weapons systems,'' \emph{arXiv preprint arXiv:2502.10174}, 2025.

\bibitem{fu2024imprompter}
X.~Fu, S.~Li, Z.~Wang, Y.~Liu, R.~K. Gupta, T.~Berg-Kirkpatrick, and E.~Fernandes, ``Imprompter: Tricking llm agents into improper tool use,'' \emph{arXiv preprint arXiv:2410.14923}, 2024.

\bibitem{wang2025tool}
B.~Wang, X.~Chen, Z.~Wang, and C.~Xiao, ``Tool-induced misalignment in llm agents: A causal analysis,'' in \emph{International Conference on Learning Representations}, 2025.

\bibitem{domkundwar2024safeguarding}
I.~Domkundwar, M.~N~S, and I.~Bhola, ``Safeguarding ai agents: Developing and analyzing safety architectures,'' \emph{arXiv preprint arXiv:2409.03793}, 2024.

\bibitem{morris2023levels}
M.~R. Morris, J.~Sohl-Dickstein, N.~Fiedel, T.~Warkentin, A.~Dafoe, A.~Faust, C.~Farabet, and S.~Legg, ``Levels of {{AGI}} for operationalizing progress on the path to {{AGI}},'' \emph{arXiv preprint arXiv:2311.02462}, 2023.

\bibitem{cihon2025measuring}
P.~Cihon, M.~Stein, G.~Bansal, S.~Manning, and K.~Xu, ``Measuring {{AI}} agent autonomy: Towards a scalable approach with code inspection,'' \emph{arXiv preprint arXiv:2502.15212}, 2025.

\bibitem{soderlevels}
L.~Soder, J.~Smakman, C.~Dunlop, W.~Pan, S.~Swaroop, and N.~Kolt, ``Levels of autonomy: Liability in the age of {{AI}} agents,'' in \emph{Workshop on Socially Responsible Language Modelling Research}, 2025.

\bibitem{wang2025unveiling}
B.~Wang, W.~He, P.~He, S.~Zeng, Z.~Xiang, Y.~Xing, and J.~Tang, ``Unveiling privacy risks in llm agent memory,'' \emph{arXiv preprint arXiv:2502.13172}, 2025.

\bibitem{masterman2024landscape}
T.~Masterman, A.~Freed, and L.~Weng, ``The landscape of emerging ai agent architectures for reasoning, planning, and tool calling: A survey,'' arXiv preprint arXiv:2404.11584, 2024.

\bibitem{ruan2023tptu}
J.~Ruan, Y.~Chen, B.~Zhang, Z.~Xu, T.~Bao, H.~Mao, Z.~Li, X.~Zeng, R.~Zhao \emph{et~al.}, ``Tptu: Task planning and tool usage of large language model-based ai agents,'' in \emph{NeurIPS 2023 Foundation Models for Decision Making Workshop}, 2023.

\bibitem{bran2023chemcrow}
A.~M. Bran, S.~Cox, O.~Schilter, C.~Baldassari, A.~D. White, and P.~Schwaller, ``Chemcrow: Augmenting large-language models with chemistry tools,'' \emph{arXiv preprint arXiv:2304.05376}, 2023.

\bibitem{ke2025dwim}
F.~Ke, X.~Leng, Z.~Cai, Z.~Khan, W.~Wang, P.~D. Haghighi, H.~Rezatofighi, M.~Chandraker \emph{et~al.}, ``Dwim: Towards tool-aware visual reasoning via discrepancy-aware workflow generation \& instruct-masking tuning,'' \emph{arXiv preprint arXiv:2503.19263}, 2025.

\bibitem{zhang2024agentpro}
W.~Zhang, K.~Tang, H.~Wu, M.~Wang, Y.~Shen, G.~Hou, Z.~Tan, P.~Li, Y.~Zhuang, and W.~Lu, ``Agent-pro: Learning to evolve via policy-level reflection and optimization,'' \emph{arXiv preprint arXiv:2402.17574}, 2024.

\bibitem{putta2024agent}
P.~Putta, E.~Mills, N.~Garg, S.~Motwani, C.~Finn, D.~Garg, and R.~Rafailov, ``Agent q: Advanced reasoning and learning for autonomous ai agents,'' \emph{arXiv preprint arXiv:2408.07199}, 2024.

\bibitem{perez2023ignore}
L.~Perez and O.~K. Hossain, ``Ignore previous prompt: Attack techniques for language models,'' arXiv preprint arXiv:2304.09701, 2023.

\bibitem{zou2023universal}
J.~Zou, T.~Xiao, R.~He \emph{et~al.}, ``Universal and transferable adversarial attacks on aligned language models,'' arXiv preprint arXiv:2307.15043, 2023.

\bibitem{autogpt2023}
S.~Gravitas, ``Auto-gpt: An experimental open-source attempt to make gpt-4 fully autonomous,'' 2023.

\bibitem{shinn2023reflexion}
N.~Shinn, F.~Cassano, A.~Gopinath, K.~Narasimhan, and S.~Yao, ``Reflexion: Language agents with verbal reinforcement learning,'' \emph{Advances in Neural Information Processing Systems}, vol.~36, pp. 8634--8652, 2023.

\bibitem{bengio2023world}
Y.~Bengio, A.~Goyal, and M.~Halawa, ``World models and active inference,'' arXiv preprint arXiv:2302.06696, 2023.

\bibitem{kirsch2021rlwm}
L.~Kirsch, E.~Grefenstette, and M.~Botvinick, ``Reinforcement learning with learned world models,'' in \emph{Advances in Neural Information Processing Systems (NeurIPS)}, 2021.

\bibitem{yudkowsky2023agi}
E.~Yudkowsky, ``Agi ruin: A list of lethalities,'' 2023.

\bibitem{christiano2017deep}
P.~F. Christiano, J.~Leike, T.~Brown \emph{et~al.}, ``Deep reinforcement learning from human preferences,'' in \emph{Advances in Neural Information Processing Systems (NeurIPS)}, 2017.

\bibitem{kumar2024refusal}
P.~Kumar, E.~Lau, S.~Vijayakumar, T.~Trinh, S.~R. Team, E.~Chang, V.~Robinson, S.~Hendryx, S.~Zhou, M.~Fredrikson \emph{et~al.}, ``Refusal-trained llms are easily jailbroken as browser agents,'' \emph{arXiv preprint arXiv:2410.13886}, 2024.

\bibitem{yuan2024r}
T.~Yuan, Z.~He, L.~Dong, Y.~Wang, R.~Zhao, T.~Xia, L.~Xu, B.~Zhou, F.~Li, Z.~Zhang \emph{et~al.}, ``R-judge: Benchmarking safety risk awareness for llm agents,'' \emph{arXiv preprint arXiv:2401.10019}, 2024.

\bibitem{goodyear2025effect}
L.~Goodyear, R.~Guo, and R.~Johari, ``The effect of state representation on llm agent behavior in dynamic routing games,'' \emph{arXiv preprint arXiv:2506.15624}, 2025.

\bibitem{zhang2025agentalign}
J.~Zhang, L.~Yin, Y.~Zhou, and S.~Hu, ``Agentalign: Navigating safety alignment in the shift from informative to agentic large language models,'' \emph{arXiv preprint arXiv:2505.23020}, 2025.

\bibitem{kim2025medical}
Y.~Kim, H.~Jeong, S.~Chen, S.~S. Li, M.~Lu, K.~Alhamoud, J.~Mun, C.~Grau, M.~Jung, R.~R. Gameiro, L.~Fan, E.~Park, T.~Lin, J.~Yoon, W.~Yoon, M.~Sap, Y.~Tsvetkov, P.~P. Liang, X.~Xu, X.~Liu, D.~McDuff, H.~Lee, H.~W. Park, S.~R. Tulebaev, and C.~Breazeal, ``Medical hallucinations in foundation models and their impact on healthcare,'' \emph{medRxiv}, vol. 2025.02.28.25323115, 2025.

\bibitem{ziegler2019finetuning}
D.~M. Ziegler, N.~Stiennon, J.~Wu, T.~B. Brown, A.~Radford, D.~Amodei, P.~Christiano, and G.~Irving, ``Fine-tuning language models from human preferences,'' \emph{arXiv preprint arXiv:1909.08593}, 2019.

\bibitem{stiennon2020learning}
N.~Stiennon, L.~Ouyang, J.~Wu, D.~Ziegler, R.~Lowe, C.~Voss, A.~Radford, D.~Amodei, and P.~F. Christiano, ``Learning to summarize with human feedback,'' \emph{Advances in neural information processing systems}, vol.~33, pp. 3008--3021, 2020.

\bibitem{ouyang2022training}
L.~Ouyang, J.~Wu, X.~Jiang, D.~Almeida, C.~Wainwright, P.~Mishkin, C.~Zhang, S.~Agarwal, K.~Slama, A.~Ray \emph{et~al.}, ``Training language models to follow instructions with human feedback,'' \emph{Advances in neural information processing systems}, vol.~35, pp. 27\,730--27\,744, 2022.

\bibitem{bondarenko2025demonstrating}
A.~Bondarenko, D.~Volk, D.~Volkov, and J.~Ladish, ``Demonstrating specification gaming in reasoning models,'' \emph{arXiv preprint arXiv:2502.13295}, 2025.

\bibitem{he2024emerged}
F.~He, T.~Zhu, D.~Ye, B.~Liu, W.~Zhou, and P.~S. Yu, ``The emerged security and privacy of llm agent: A survey with case studies,'' \emph{arXiv preprint arXiv:2407.19354}, 2024.

\bibitem{amodei2016concrete}
D.~Amodei, C.~Olah, J.~Steinhardt, P.~Christiano, J.~Schulman, and D.~Man{\'e}, ``Concrete problems in {{AI}} safety,'' \emph{arXiv preprint arXiv:1606.06565}, 2016.

\bibitem{guo2023empowering}
J.~Guo, N.~Li, J.~Qi, H.~Yang, R.~Li, Y.~Feng, S.~Zhang, and M.~Xu, ``Empowering working memory for large language model agents,'' \emph{arXiv preprint arXiv:2312.17259}, 2023.

\bibitem{openai2023redteaming}
OpenAI, ``{{OpenAI}} red teaming network,'' OpenAI, Tech. Rep., 2023.

\bibitem{chen2022redteaming}
A.~Chen \emph{et~al.}, ``A survey of red teaming in nlp: Attacks, evaluation, and defense,'' \emph{arXiv preprint arXiv:2210.09819}, 2022.

\bibitem{perez2022ignore}
F.~Perez and I.~Ribeiro, ``Ignore previous prompt: Attack techniques for language models,'' \emph{arXiv preprint arXiv:2211.09527}, 2022.

\bibitem{mo2024trembling}
L.~Mo, Z.~Liao, B.~Zheng, Y.~Su, C.~Xiao, and H.~Sun, ``A trembling house of cards? mapping adversarial attacks against language agents,'' \emph{arXiv preprint arXiv:2402.10196}, 2024.

\bibitem{wang2024badagent}
Y.~Wang, D.~Xue, S.~Zhang, and S.~Qian, ``Badagent: Inserting and activating backdoor attacks in llm agents,'' \emph{arXiv preprint arXiv:2406.03007}, 2024.

\bibitem{yuan2019adversarial}
X.~Yuan, P.~He, Q.~Zhu, and X.~Li, ``Adversarial examples: Attacks and defenses for deep learning,'' \emph{IEEE Transactions on Neural Networks and Learning Systems}, vol.~30, no.~9, pp. 2805--2824, 2019.

\bibitem{andriushchenko2024agentharm}
M.~Andriushchenko, A.~Souly, M.~Dziemian, D.~Duenas, M.~Lin, J.~Wang, D.~Hendrycks, A.~Zou, Z.~Kolter, M.~Fredrikson \emph{et~al.}, ``Agentharm: A benchmark for measuring harmfulness of llm agents,'' \emph{arXiv preprint arXiv:2410.09024}, 2024.

\bibitem{zhang2024multitrust}
Y.~Zhang, Y.~Huang, Y.~Sun, C.~Liu, Z.~Zhao, Z.~Fang, Y.~Wang, H.~Chen, X.~Yang, X.~Wei, H.~Su, Y.~Dong, and J.~Zhu, ``Multitrust: A comprehensive benchmark towards trustworthy multimodal large language models,'' in \emph{Advances in Neural Information Processing Systems}, vol.~37, 2024, pp. 49\,279--49\,383.

\bibitem{wang2023sensorattack}
H.~Wang \emph{et~al.}, ``Sensor attack taxonomy and defense for cyber-physical systems,'' \emph{ACM Computing Surveys}, 2023.

\bibitem{zhou2023scav}
Y.~Zhou \emph{et~al.}, ``Scav: Steering llm hallucination via self-checking augmented verification,'' \emph{arXiv preprint arXiv:2309.00247}, 2023.

\bibitem{carlini2021extracting}
N.~Carlini \emph{et~al.}, ``Extracting training data from large language models,'' \emph{USENIX Security Symposium}, 2021.

\bibitem{stadfeld2022membership}
C.~Stadfeld \emph{et~al.}, ``Membership inference attacks on sequence-to-sequence models,'' \emph{Proceedings of the ACM Conference on Computer and Communications Security (CCS)}, 2022.

\bibitem{kessler2022surprising}
S.~Kessler, P.~Mi{\l}o{\'s}, J.~Parker-Holder, and S.~J. Roberts, ``The surprising effectiveness of latent world models for continual reinforcement learning,'' in \emph{Deep Reinforcement Learning Workshop NeurIPS 2022}, 2022.

\bibitem{peng2023instruction}
B.~Peng \emph{et~al.}, ``Instruction tuning with gpt-4,'' \emph{arXiv preprint arXiv:2304.03277}, 2023.

\bibitem{ganguli2022redteaming}
D.~Ganguli \emph{et~al.}, ``Red teaming language models to reduce harms: Methods, scaling behaviors, and lessons learned,'' \emph{arXiv preprint arXiv:2209.07858}, 2022.

\bibitem{huang2023retrieval}
A.~Huang \emph{et~al.}, ``Retrieval-augmented language models expose sensitive information,'' \emph{arXiv preprint arXiv:2310.01819}, 2023.

\bibitem{liu2023houyi}
Y.~Liu, G.~Deng, Y.~Li, K.~Wang, Z.~Wang, X.~Wang, T.~Zhang, Y.~Liu, H.~Wang, Y.~Zheng, and Y.~Liu, ``Prompt injection attack against {LLM}-integrated applications,'' \emph{arXiv preprint arXiv:2306.05499}, 2023.

\bibitem{yi2025bipia_kdd}
J.~Yi, Y.~Xie, B.~Zhu, E.~Kiciman, G.~Sun, X.~Xie, and F.~Wu, ``Benchmarking and defending against indirect prompt injection attacks on large language models,'' in \emph{Proceedings of the 31st ACM SIGKDD Conference on Knowledge Discovery and Data Mining (KDD ’25)}, 2025.

\bibitem{huang2025breaking}
Y.~Huang, Y.~Sun, S.~Ruan, Y.~Zhang, Y.~Dong, and X.~Wei, ``Breaking the ceiling: Exploring the potential of jailbreak attacks through expanding strategy space,'' \emph{arXiv preprint arXiv:2505.21277}, 2025.

\bibitem{hung2025attention}
K.-H. Hung, C.-Y. Ko, A.~Rawat, I.-H. Chung, W.~H. Hsu, and P.-Y. Chen, ``Attention tracker: Detecting prompt injection attacks in {LLM}s,'' in \emph{Findings of the Association for Computational Linguistics: NAACL 2025}, 2025, pp. 2309--2322.

\bibitem{pathade2025redteaming}
C.~Pathade, ``Red teaming the mind of the machine: A systematic evaluation of prompt injection and jailbreak vulnerabilities in llms,'' \emph{arXiv preprint arXiv:2505.04806}, 2025.

\bibitem{dong2023robust}
Y.~Dong, H.~Chen, J.~Chen, Z.~Fang, X.~Yang, Y.~Zhang, Y.~Tian, H.~Su, and J.~Zhu, ``How robust is google's bard to adversarial image attacks?'' in \emph{R0-FoMo Workshop on Robustness of Few-shot and Zero-shot Learning in Large Foundation Models in Advances in Neural Information Processing Systems}, 2023.

\bibitem{wu2025adversarial_agents}
C.~H. Wu, R.~R. Shah, J.~Y. Koh, R.~Salakhutdinov, D.~Fried, and A.~Raghunathan, ``Adversarial attacks on multimodal agents,'' in \emph{Proceedings of the International Conference on Learning Representations (ICLR 2025)}, 2025.

\bibitem{guan2024jmtfa}
J.~Guan, T.~Ding, L.~Cao, L.~Pan, C.~Wang, and X.~Zheng, ``Probing the robustness of vision--language pretrained models: A multimodal adversarial attack approach,'' \emph{arXiv preprint arXiv:2406.13461}, 2024.

\bibitem{zhou2025guardian}
J.~Zhou, L.~Wang, and X.~Yang, ``Guardian: Safeguarding llm multi-agent collaborations with temporal graph modeling,'' \emph{arXiv preprint arXiv:2505.19234}, 2025.

\bibitem{owasp2025prompt}
{OWASP Gen AI Security Project}, ``Llm01:2025 prompt injection,'' Online technical report, 2025.

\bibitem{kierans2024quantifying}
A.~Kierans, H.~Hazan, and S.~Dori-Hacohen, ``Quantifying misalignment between agents: Towards a sociotechnical understanding of alignment,'' \emph{arXiv preprint arXiv:2406.04231}, 2024.

\bibitem{dong2025practical}
S.~Dong, S.~Xu, P.~He, Y.~Li, J.~Tang, T.~Liu, H.~Liu, and Z.~Xiang, ``A practical memory injection attack against llm agents,'' \emph{arXiv preprint arXiv:2503.03704}, 2025.

\bibitem{Hatalis2024Memory}
K.~Hatalis, D.~Christou, J.~Myers, S.~Jones, K.~Lambert, A.~Amos-Binks, Z.~Dannenhauer, and D.~Dannenhauer, ``Memory matters: The need to improve long-term memory in llm-agents,'' \emph{Proceedings of the AAAI Symposium Series}, vol.~2, no.~1, pp. 277--280, 2024.

\bibitem{Bostrom2014}
N.~Bostrom, \emph{Superintelligence: Paths, Dangers, Strategies}.\hskip 1em plus 0.5em minus 0.4em\relax Oxford University Press, 2014.

\bibitem{zhang2024agentsafetybench}
Z.~Zhang, S.~Cui, Y.~Lu, J.~Zhou, J.~Yang, H.~Wang, and M.~Huang, ``Agent-safetybench: Evaluating the safety of llm agents,'' \emph{arXiv preprint arXiv:2412.14470}, 2024.

\bibitem{ge2023llm}
Y.~Ge, Y.~Ren, W.~Hua, S.~Xu, J.~Tan, and Y.~Zhang, ``Llm as os, agents as apps: Envisioning aios, agents and the aios-agent ecosystem,'' \emph{arXiv preprint arXiv:2312.03815}, 2023.

\bibitem{brown2020language}
T.~B. Brown, B.~Mann, N.~Ryder, M.~Subbiah, and J.~Kaplan, ``Language models are few-shot learners,'' in \emph{Proceedings of NeurIPS 2020}, vol.~33, 2020, pp. 1877--1901.

\bibitem{jilk2018limits}
D.~J. Jilk, ``Limits to verification and validation of agentic behavior,'' in \emph{Artificial Intelligence Safety and Security}, 2018, pp. 225--234.

\bibitem{Fang2024Teams}
R.~Fang, R.~Bindu, A.~Gupta, Q.~Zhan, and D.~Kang, ``Teams of {{LLM}} agents can exploit zero-day vulnerabilities,'' \emph{arXiv preprint arXiv:2406.01637}, 2024.

\bibitem{liu2025cascade}
Y.~Liu, Y.~Zhang, X.~Wang, and D.~Chen, ``Cascading hallucination in multi-agent systems,'' \emph{Nature Machine Intelligence}, vol.~7, no.~3, pp. 145--156, 2025.

\bibitem{zhang2024lamas}
Y.~Zhang, L.~Zhou, T.~Schick, and J.~Lin, ``Techniques and business perspectives llm-based multi-agent systems,'' \emph{arXiv preprint arXiv:2411.12076}, 2024.

\bibitem{He2025Red-Teaming}
P.~He, Y.~Lin, S.~Dong, H.~Xu, Y.~Xing, and H.~Liu, ``Red-teaming llm multi-agent systems via communication attacks,'' \emph{arXiv preprint arXiv:2502.14847}, 2025.

\bibitem{altman1999constrained}
E.~Altman, \emph{Constrained Markov Decision Processes}.\hskip 1em plus 0.5em minus 0.4em\relax CRC Press, 1999.

\bibitem{kushwaha2025survey}
A.~Kushwaha, K.~Ravish, P.~Lamba, and P.~Kumar, ``A survey of safe reinforcement learning and constrained mdps: Foundations and extensions,'' \emph{arXiv preprint arXiv:2505.17342}, 2025.

\bibitem{achiam2017constrained}
J.~Achiam, D.~Held, A.~Tamar, and P.~Abbeel, ``Constrained policy optimization,'' in \emph{Proceedings of the 34th International Conference on Machine Learning (ICML)}, vol.~70, 2017, pp. 22--31.

\bibitem{ray2019benchmarking}
A.~Ray, J.~Achiam, and D.~Amodei, ``Benchmarking safe exploration in deep reinforcement learning,'' \emph{arXiv preprint arXiv:1910.01708}, 2019.

\bibitem{chen2025agentpoison}
Z.~Chen, Z.~Xiang, C.~Xiao, D.~Song, and B.~Li, ``Agentpoison: Red‑teaming llm agents via poisoning memory or knowledge bases,'' \emph{NeurIPS (Poster)}, 2024.

\bibitem{fu2024imprompt}
X.~Fu, Y.~Song, and P.~Liu, ``Imprompt: Prompt-based attacks on tool-using llm agents,'' \emph{arXiv preprint arXiv:2410.14923}, 2024.

\bibitem{Lee2024PromptInfection}
D.~Lee and M.~Tiwari, ``Prompt infection: {{LLM-to-LLM}} prompt injection within multi-agent systems,'' \emph{arXiv preprint arXiv:2410.07283}, 2024.

\bibitem{Yang2024Backdoor}
W.~Yang, X.~Bi, Y.~Lin, S.~Chen, J.~Zhou, and X.~Sun, ``Watch out for your agents! investigating backdoor threats to {{LLM}}-based agents,'' \emph{arXiv preprint arXiv:2402.11208}, 2024.

\bibitem{Nakash2024ReActAgents}
I.~Nakash, G.~Kour, G.~Uziel, and A.~Anaby-Tavor, ``Breaking react agents: Foot-in-the-door attack will get you in,'' \emph{arXiv preprint arXiv:2410.16950}, 2024.

\bibitem{Triedman2025MultiAgent}
H.~Triedman, R.~Jha, and V.~Shmatikov, ``Multi-agent systems execute arbitrary malicious code,'' \emph{arXiv preprint arXiv:2503.12188}, 2025.

\bibitem{luo2025agentauditor}
H.~Luo, S.~Dai, C.~Ni, X.~Li, G.~Zhang, K.~Wang, T.~Liu, and H.~Salam, ``Agentauditor: Human-level safety and security evaluation for llm agents,'' \emph{arXiv preprint arXiv:2506.00641}, 2025.

\bibitem{Pan2022RewardMisspecification}
A.~Pan, K.~Bhatia, and J.~Steinhardt, ``The effects of reward misspecification: Mapping and mitigating misaligned models,'' \emph{arXiv preprint arXiv:2201.03544}, 2022.

\bibitem{Skalse2022DefiningHacking}
J.~Skalse, N.~Howe, D.~Krasheninnikov, and D.~Krueger, ``Defining and characterizing reward hacking,'' \emph{arXiv preprint arXiv:2209.13085}, 2022.

\bibitem{Weng2024RewardHacking}
L.~Weng, ``Reward hacking in reinforcement learning,'' \emph{Lilian’s ML Blog}, 2024.

\bibitem{Farquhar2025MONA}
S.~Farquhar, V.~Varma, and D.~e.~a. Lindner, ``Mona: Myopic optimization with non-myopic approval can mitigate multi-step reward hacking,'' \emph{arXiv preprint arXiv:2501.13011}, 2025.

\bibitem{deng2024ai}
Z.~Deng, Y.~Guo, C.~Han, W.~Ma, J.~Xiong, S.~Wen, and Y.~Xiang, ``Ai agents under threat: A survey of key security challenges and future pathways,'' \emph{arXiv preprint arXiv:2406.02630}, 2024.

\bibitem{Yoffe2024DebUnc}
L.~Yoffe, A.~Amayuelas, and W.~Y. Wang, ``Debunc: Mitigating hallucinations in large language model agent communication with uncertainty estimations,'' \emph{arXiv preprint arXiv:2407.06426}, 2024.

\bibitem{He2025AiTM}
P.~He, Y.~Lin, S.~Dong, H.~Xu, Y.~Xing, and H.~Liu, ``Red-teaming llm multi-agent systems via communication attacks,'' \emph{arXiv preprint arXiv:2502.14847}, 2025.

\bibitem{tian2023evil}
Y.~Tian, X.~Yang, J.~Zhang, Y.~Dong, and H.~Su, ``Evil geniuses: Delving into the safety of llm-based agents,'' \emph{arXiv preprint arXiv:2311.11855}, 2023.

\bibitem{solomonoff1964formal}
R.~J. Solomonoff, ``A formal theory of inductive inference. part i,'' \emph{Information and Control}, vol.~7, no.~1, pp. 1--22, 1964.

\bibitem{soares2015formal}
N.~Soares and B.~Fallenstein, ``Formalizing two problems of realistic world models,'' 2015.

\bibitem{soares2015logical}
N.~Soares, B.~Fallenstein, P.~Christiano, and J.~Taylor, ``Questions of reasoning under logical uncertainty,'' \emph{Technical report}, 2015.

\bibitem{lin2023vila}
J.~Lin, H.~Yin, W.~Ping, Y.~Lu, P.~Molchanov, A.~Tao, H.~Mao, J.~Kautz, M.~Shoeybi, and S.~Han, ``Vila: On pre-training for visual language models,'' \emph{arXiv preprint arXiv:2312.07533}, 2023.

\bibitem{chen2024MAT}
X.~Chen, H.~Li, R.~Zhang, and M.~Wang, ``Mat: Multimodal adversarial training for vision–language models,'' \emph{arXiv preprint arXiv:2405.18770}, 2024.

\bibitem{chen2024struq}
S.~Chen, J.~Piet, C.~Sitawarin, and D.~Wagner, ``Struq: Defending against prompt injection with structured queries,'' \emph{arXiv preprint arXiv:2402.06363}, 2024.

\bibitem{lee2024prompt}
D.~Lee and M.~Tiwari, ``Prompt infection: {{LLM-to-LLM}} prompt injection within multi-agent systems,'' \emph{arXiv preprint arXiv:2410.07283}, 2024.

\bibitem{cobb2022context}
O.~Cobb and A.~Van~Looveren, ``Context-aware drift detection,'' in \emph{International Conference on Machine Learning}.\hskip 1em plus 0.5em minus 0.4em\relax PMLR, 2022, pp. 4087--4111.

\bibitem{chen2024riskTransformer}
J.~Zhang, H.~Xie, X.~Zhang, and K.~Liu, ``Enhancing risk assessment in transformers with loss-at-risk functions,'' \emph{arXiv preprint arXiv:2411.02558}, 2024.

\bibitem{wu2023edt}
Y.-H. Wu, X.~Wang, and M.~Hamaya, ``Elastic decision transformer,'' in \emph{Proceedings of the 2023 International Conference on Learning Representations (ICLR)}, 2023.

\bibitem{tang2024ardt}
X.~Tang, A.~Marques, P.~Kamalaruban, and I.~Bogunovic, ``Adversarially robust decision transformer,'' in \emph{Proceedings of the 2024 International Conference on Learning Representations (ICLR)}, 2024.

\bibitem{hubinger2019risks}
E.~Hubinger and et~al., ``Risks from learned optimization in advanced machine learning systems,'' \emph{Technical Report, MIRI}, 2019.

\bibitem{luo2024processSupervision}
L.~Luo, Y.~Liu \emph{et~al.}, ``Process supervision: Aligning language models via intermediates,'' \emph{arXiv preprint arXiv:2406.06592}, 2024.

\bibitem{qin2024toolllm}
Y.~Qin, S.~Liang, Y.~Ye, K.~Zhu, L.~Yan, Y.~Lu, Y.~Lin, X.~Cong, X.~Tang, B.~Qian, S.~Zhao, L.~Hong, R.~Tian, R.~Xie, J.~Zhou, M.~Gerstein, D.~Li, Z.~Liu, and M.~Sun, ``Toolllm: Facilitating large language models to master 16,000+ real‑world apis,'' in \emph{The 12th International Conference on Learning Representations (ICLR)}, 2024.

\bibitem{zheng2025calm}
X.~Zheng, L.~Wang, Y.~Liu, X.~Ma, C.~Shen, and C.~Wang, ``Calm: Curiosity‑driven auditing for large language models,'' in \emph{Proceedings of the AAAI Conference on Artificial Intelligence}, vol.~39, no.~26, 2025, pp. 27\,757--27\,764.

\bibitem{zhou2023rlaif}
K.~Zhou \emph{et~al.}, ``Rlaif: Reinforcement learning from human feedback to mitigate reward hacking,'' in \emph{OpenReview (NeurIPS Workshop)}, 2023.

\bibitem{uncknown2024faultHealing}
Unknown, ``Fault self‑healing: A biological immune heuristic reinforcement approach,'' \emph{Engineering Applications of Artificial Intelligence}, 2024.

\bibitem{gao2024prelude}
T.~Gao, A.~Chen, B.~Zhang, J.~Liu, P.~Li, P.~Liang, and T.~Hashimoto, ``Aligning llm agents by learning latent preferences from user edits (prelude),'' in \emph{Proceedings of the 38th Conference on Neural Information Processing Systems (NeurIPS)}, 2024.

\bibitem{schlarmann2024robustCLIP}
C.~Schlarmann, N.~Singh, F.~Croce, and M.~Hein, ``Robust clip: Unsupervised adversarial fine-tuning of vision embeddings,'' \emph{arXiv preprint arXiv:2402.12336}, 2024.

\bibitem{chen2025evaluating}
Y.~Chen, X.~Hu, K.~Yin, J.~Li, and S.~Zhang, ``Evaluating the robustness of multimodal agents against active environmental injection attacks,'' \emph{arXiv preprint arXiv:2502.13053}, 2025.

\bibitem{jiang2025survey}
C.~Jiang, Z.~Wang, M.~Dong, and J.~Gui, ``Survey of adversarial robustness in multimodal large language models,'' \emph{arXiv preprint arXiv:2503.13962}, 2025.

\bibitem{xu2025amem}
M.~Xu, T.~Liang, S.~Mei, and Z.~Huang, ``A-mem: Agentic memory for llm agents,'' \emph{arXiv preprint arXiv:2502.12110}, 2025.

\bibitem{cheng2024mini}
H.~Cheng, M.~Zhang, and J.~Q. Shi, ``Mini-llm: Memory-efficient structured pruning for large language models,'' \emph{arXiv preprint arXiv:2407.11681}, 2024.

\bibitem{wu2024multimodalRobustness}
H.~C. Wu, R.~Shah, J.~Y. Koh, R.~Salakhutdinov, D.~Fried, and A.~Raghunathan, ``Dissecting adversarial robustness of multimodal lm agents,'' \emph{arXiv preprint arXiv:2406.12814}, 2024.

\bibitem{amirizaniani2024auditllm}
M.~Amirizaniani, E.~Martin, T.~Roosta, A.~Chadha, and C.~Shah, ``Auditllm: a tool for auditing large language models using multiprobe approach,'' in \emph{Proceedings of the 33rd ACM International Conference on Information and Knowledge Management}, 2024, pp. 5174--5179.

\bibitem{sun2023riskq}
W.~Sun, Q.~Zhao, and X.~Chen, ``Riskq: Risk-sensitive multi-agent reinforcement learning value factorization,'' in \emph{Advances in Neural Information Processing Systems (NeurIPS)}, 2023.

\bibitem{zhou2025self}
Y.~Zhou, S.~Levine, J.~Weston, X.~Li, and S.~Sukhbaatar, ``Self-challenging language model agents,'' \emph{arXiv preprint arXiv:2506.01716}, 2025.

\bibitem{qin2023toolllm}
Y.~Qin, S.~Liang, Y.~Ye, K.~Zhu, L.~Yan, Y.~Lu, Y.~Lin, X.~Cong, X.~Tang, B.~Qian \emph{et~al.}, ``Toolllm: Facilitating large language models to master 16000+ real-world apis,'' \emph{arXiv preprint arXiv:2307.16789}, 2023.

\bibitem{paternain2019constrained}
S.~Paternain, K.~Shanmugam, A.~Krause, and A.~Ribeiro, ``Constrained reinforcement learning has zero duality gap,'' in \emph{Advances in Neural Information Processing Systems}, vol.~32, 2019.

\bibitem{weng2023safe}
L.~Weng, ``Llm powered autonomous agents,'' 2023.

\bibitem{shieldagent2025}
Z.~Chen, M.~Kang, and B.~Li, ``Shieldagent: Shielding agents via verifiable safety policy reasoning,'' \emph{arXiv preprint arXiv:2503.22738}, 2025.

\bibitem{zhang2024fusion}
Y.~Zhang, Y.~Cai, X.~Zuo, X.~Luan, K.~Wang, Z.~Hou, Y.~Zhang, Z.~Wei, M.~Sun, J.~Sun, J.~Sun, and J.~S. Dong, ``The fusion of large language models and formal methods for trustworthy ai agents: A roadmap,'' \emph{arXiv preprint arXiv:2412.06512}, 2024.

\bibitem{zhang2025stair}
Y.~Zhang, S.~Zhang, Y.~Huang, Z.~Xia, Z.~Fang, X.~Yang, R.~Duan, D.~Yan, Y.~Dong, and J.~Zhu, ``Stair: Improving safety alignment with introspective reasoning,'' \emph{arXiv preprint arXiv:2502.02384}, 2025.

\bibitem{zhang2025realsafe}
Y.~Zhang, Z.~Zeng, D.~Li, Y.~Huang, Z.~Deng, and Y.~Dong, ``Realsafe-r1: Safety-aligned deepseek-r1 without compromising reasoning capability,'' \emph{arXiv preprint arXiv:2504.10081}, 2025.

\bibitem{rerefst2024}
Z.~Dou, C.~Yang, X.~Wu, K.~Chang, and N.~Peng, ``Re‑rest: Reflection‑reinforced self‑training for language agents,'' \emph{arXiv preprint arXiv:2406.01495}, 2024.

\bibitem{aegisllm2025}
Z.~Cai, S.~Shabihi, B.~An, Z.~Che, B.~R. Bartoldson, B.~Kailkhura, T.~Goldstein, and F.~Huang, ``Aegisllm: Scaling agentic systems for self‑reflective defense in llm security,'' \emph{arXiv preprint arXiv:2504.20965}, 2025.

\bibitem{copper2024}
X.~Bo, Z.~Zhang, Q.~Dai, X.~Feng, L.~Wang, R.~Li, X.~Chen, and J.~Wen, ``Reflective multi‑agent collaboration based on large language models,'' in \emph{Advances in Neural Information Processing Systems 37}, 2024, pp. 13\,859--13\,875.

\bibitem{maebe2025}
S.~Erisken, T.~Gothard, M.~Leitgab, and R.~Potham, ``Maebe: Multi-agent emergent behavior evaluation framework,'' \emph{arXiv preprint arXiv:2506.03053}, 2025.

\end{thebibliography}
}

\end{document}